# Learning the Topic, Not the Language: How LLMs Classify Online Immigration Discourse Across Languages

Andrea Nasuto[1,2], Stefano Iacus[1], Francisco Rowe[2], and Devika Jain[1]

[1]Harvard University
[2]University of Liverpool

**Abstract**

Large language models (LLMs) offer new opportunities for scalable analysis of online discourse. Yet their use in multilingual social science research remains constrained by model size, cost and linguistic bias. We develop a lightweight, open-source LLM framework using fine-tuned LLaMA 3.2–3B models to classify immigration-related tweets across 13 languages. Unlike prior work relying on BERT-style models or translation pipelines, we combine topic classification with stance detection and demonstrate that LLMs fine-tuned in just one or two languages can generalize topic understanding to unseen languages. Capturing ideological nuance, however, benefits from multilingual fine-tuning. Our approach corrects pretraining biases with minimal data from under-represented languages and avoids reliance on proprietary systems. With 26–168× faster inference and over 1000× cost savings compared to commercial LLMs, our method supports real-time analysis of billions of tweets. This scale-first framework enables inclusive, reproducible research on public attitudes across linguistic and cultural contexts.

**Significance**

We show that fine-tuned LLMs in one or two languages can reliably classify a culturally rich and highly polarised topic such as public debate on migration across 11 additional languages not seen during training. Yet, capturing nuanced stances (pro, anti, neutral) requires exposure to more diverse linguistic data. Incorporating even a few thousand tokens from low-resource languages during fine-tuning mitigates the English-centric bias inherited from pretraining and substantially improves classification accuracy. By developing and deploying 4-bit-quantized, LoRA-adapted open models, we provide an open-source, reproducible alternative to proprietary LLMs with 36–136 times faster inference and over 10 million times lower cost than GPT-4o. Our approach advances theory on cross-lingual representation and delivers practical tools for inclusive multilingual research.

# 1 Introduction

The widespread availability of generative AI, particularly large language models (LLMs), is transforming scientific research and more generally human practices. These models are more accessible, easy to use and deployed across a range of text-based tasks in all academic fields. In the social sciences, LLMs have been used to analyze political speech [1], finance sentiment [2], social media opinions on immigration [3], enhance systematic literature reviews [4] and improve text classification [5].

Despite the growing adoption of LLMs, important gaps remain in our understanding of how fine tuned LLMs can classify text across multiple languages while remaining cost effective and environmentally sustainable at scale. Most prior work relies on encoder based BERT family models rather than decoder only LLMs, evaluates performance on benchmark datasets without considering the challenge of deploying models across trillions of tokens, and focuses primarily on hate speech classification instead of jointly modeling topic relevance and nuanced stances toward immigration on online social media, which has become a key societal challenge across the globe.

In this paper, we study whether a relatively small, open source decoder-only LLM can be fine tuned to classify a culturally rich and politically salient issue, such as immigration across multiple languages using different highly efficient and environmental-friendly fine-tuned LLMs. We also assess whether that model can generalize both topic and stance classification across many languages while remaining financially and computationally efficient to be deployed at scale. We go beyond legacy encoder-based BERT models, which dominate much of the earlier literature. We compare several fine tuning setups, including models trained on a single language, on two languages and on a multilingual mix, and evaluate how well each configuration transfers to unseen languages.

Over the past decade, the underlying architectures and scales of these models have changed substantially. Early transformer based systems such as BERT were encoder-only models with roughly 110M to 340M parameters, pretrained on about 3.3B tokens [6]. Legacy encoder based architectures in the BERT family, including multilingual BERT and RoBERTa, have been the workhorse of most prior cross lingual classification research because they are open source and easy to fine tune, and their relatively lightweight size also makes them appealing for large scale deployment [7, 8]. The current generation of decoder-only LLMs, including OpenAI's GPT style models and Meta's Llama family, scales both parameters and pretraining data by orders of magnitude. For example, Llama 3.2 3B has about 3B parameters, about 10 times larger than BERT, and it was pretrained on 15.6 trillion tokens [9]. This shift is not merely incremental: at these scales, models display emergent capabilities which are not predictable by extrapolating small model properties [10] including general purpose multi-lingual reasoning that can follow instructions and handle complex, context and knowledge intensive tasks [11]. Recent work suggests that fine tuned decoder only models such as LLaMA and GPT perform better than fine tuned BERT models on semantically demanding and knowledge intensive classification tasks that involve implicit meaning, coded expressions and obfuscated hostility [11], all of which strongly characterize immigration discourse on social media. At the same time, [12] show that a fine tuned LLaMA 3.1 8B model outclasses BERT based models in detecting hate speech in a culturally rich and multilingual setting with coded language. Consequently, findings and properties established in prior work on BERT family encoders should not be assumed to extend to modern instruction tuned LLMs. Taken together, these differences mean that while BERT family models remain useful points of comparison, a decoder only LLM such as LLaMA 3.2 with 3 billion parameters is a more appropriate backbone for the high context, multilingual stance classification problem we study here.

The transformative potential of LLMs does not come only from scaling up pretraining data or increasing the number of parameters, but from their ability to be adapted through fine-tuning. Fine-tuning allows turning a general-purpose foundation model into a task-specific system using domain-relevant labelled data, improving task performance and encoding domain-specific knowledge without the need to retrain a model from scratch.

Social science applications could in principle leverage these growing capabilities to study large scale text corpora, but as datasets grow to hundreds of billions or trillions of tokens, running LLMs at inference scale raises substantial computational, financial and environmental costs [REF]. Running inference with large LLMs or commercial APIs on such volumes can quickly exceed typical academic budgets and has nontrivial energy demands, which constrains the kinds of models that can be used in practice.

Existing work on fine tuned LLM classifiers has therefore tended to operate either on small curated benchmarks or synthetic tasks [13], or to rely on encoder based BERT family models that are efficient and more sustainable but underperform decoder only LLMs on multilingual, knowledge intensive and reasoning heavy tasks [11]. When decoder only models are used, they are often very large (e.g. Llama 70B) or accessed via commercial APIs such as ChatGPT, which makes it difficult to assess whether these setups can be deployed cost effectively at scale for social science workflows. Past studies typically focus on binary labels [14], are fine tuned in a single high-resource language such as English [REF], and do not test whether a single fine tuned model can jointly handle topic detection and multi class stance classification across many languages.

In multilingual classification tasks, a straightforward strategy is to fine tune a separate LLM for each language in the dataset. Yet fine tuning on many languages is costly in terms of time, computational resources and labeled data [15]). One common workaround is to translate all content into a single language, typically English, and fine tune a model only on that language. However, this approach is known to introduce translation biases that harm classification performance and can amplify cultural and gender biases [9, 16, 17, 18]. While translating with state of the art LLM based systems may be feasible for small or medium sized corpora, applying this strategy to datasets on the order of trillions of tokens would require either substantial computational infrastructure or financial resources in the millions of dollars when using commercial APIs (see Appendix (cite table with financial cost). This makes naive translate then classify pipelines impractical for large scale social media analysis.

These open questions are especially salient in the case of online public debate on immigration, a globally salient and ideologically charged topic that intersects with debates over national identity, labor markets, human rights and security [19]. Its discussion is shaped by context specific cultural and geographic dynamics, and it frequently provokes polarized public opinion, including physical violence. Social media platforms have become central arenas for immigration discourse, where anti immigration narratives, misinformation and hate speech often circulate [3]. These platforms not only reflect public attitudes but can exacerbate divisions and amplify hostility toward migrants [20]. This complex communicative landscape makes immigration an ideal testbed for evaluating how LLMs generalize their classification capabilities across languages. Yet little is known about the cross lingual properties of fine tuned LLMs in understanding such culturally rich and politically charged content.

Most existing work classifying social media content has focused on using LLM for hate speech detection [7]. These problems are typically framed as a binary decision between hateful and non hateful content. By contrast, our task requires first deciding whether a tweet is related or unrelated to immigration, and if this is the case, our model seek to classify if posts express a pro-immigration, anti-immigration or neutral stance. This narrower focus on immigration makes the task more challenging and realistic than identifying hate speech. Hate speech is a broad umbrella category that can cover misogyny, racism, homophobia and many other target concepts. Immigration expressions include subtle policy disagreements, coded language and context dependent cultural references. Although some anti-immigration messages clearly overlap with hate speech, our labels are not binary by also capturing neutral and pro-immigration stances that are typically ignored in standard hate speech classification. As such, the task of classifying immigration attitudes across languages is both a methodological challenge and a socially essential goal. Scalable, accurate classification tools are urgently needed to understand global digital discourse on immigration in real time.

Beyond hate speech, we recognize that past work has also examined the cross-lingual and cross-cultural properties of LLMs. [21] focuses on incorporating cultural differences into LLMs and measuring culture specific behavior across groups rather than multilingual classification tasks generability. [22] evaluates multilingual generative AI systems across a wide range of languages and tasks but without fine-tuning and closed-access commercial models. [23] use LLMs for cross lingual data augmentation in commonsense reasoning tasks not classification. [13] fine-tuned model to classify immigration-related content on a benchmark and only for Spanish and based on encoder models like multilanguage BERT and BETO. We are not aware of any study examining the multilingual properties of fine-tuned LLMs tested in a real case scenario with hundreds of billions of tokens to be classified, not a small benchmark dataset.

To test how well language models generalize across languages under these constraints, we fine tune the lightweight open source Meta LLaMA 3.2 model with 3 billion parameters to classify immigration related tweets using real, human annotated data. The human annotated dataset is a sample of geolocated tweets in 12 different languages and geographies extracted from the Harvard GeoTweet Archive 2.0 [24], annotated using a standardized codebook. We split the dataset into training and test sets: the training set is used to fine tune our models, and the test set is used to evaluate performance on unseen data. We experiment with 4 fine tuned models: (i) a monolingual model trained on English data only; (ii) a monolingual model trained on Spanish data only; (iii) a bilingual model trained on a combined dataset that includes both the English and Spanish data used in models 1 and 2; and (iv) a multilingual model trained on this same English Spanish dataset along with human annotated tweets in 9 additional languages, covering a total of 11 training languages and 13 evaluation languages. This experimental setup allows us to assess whether exposure to multiple languages during fine tuning leads to higher classification accuracy or if comparable results can be achieved using fewer languages. We evaluate each model on languages that were excluded from its training set, focusing on a 4 way classification task: determining whether a tweet is (i) unrelated to immigration, or expresses a (ii) neutral, (iii) pro immigration or (iv) anti immigration stance. To test the effectiveness of translation as an alternative strategy, we machine translated and then classified English translations of non English tweets. All models are trained on a near zero emission GPU cluster at Harvard and deployed in quantized format for computational efficiency. This design enables us to assess whether task specific fine tuning on one or more languages can teach LLMs to classify content in other unseen languages, particularly those underrepresented in the LLM pretraining, while remaining viable at the scale required by the GeoTweet Archive.

Our findings show that a bilingual model fine tuned in English and Spanish can reliably detect whether a tweet is about immigration, even when the tweet is written in a language the model has never seen during fine tuning. This suggests that the model learns a general representation of the topic that transfers across languages. However, the multilingual model trained on 12 languages performs better when the task requires identifying stance, that is whether a tweet expresses a pro immigration, anti immigration or neutral position. Exposure to diverse linguistic and cultural contexts during fine tuning appears to enhance the LLM's ability to capture nuanced ideological content. Importantly, these results are achieved without relying on high cost machine translation, synthetic prompts or complex multilingual pipelines. In contrast, prior studies often construct multilingual datasets by machine translating entire datasets [25], mix multiple languages within prompts [26], or chain translations of both prompts and responses [27]. Our method uses real, human annotated data and standard supervised fine tuning. This design provides a transparent and reproducible alternative that is more practical for real world multilingual applications.

To demonstrate the scalability of our approach, we estimate that our fine tuned model could classify the entire Harvard GeoTweet Archive 2.0, which contains more than 10 billion tweets, for about 2,900 dollars with virtually zero additional carbon emissions, compared to proprietary alternatives that would cost between 316,000 and 17.7 million dollars while generating hundreds to thousands of tonnes of $CO_2$. Moreover, our model operates at 3,854 tokens per second, between 36 and 168 times faster than comparable commercial alternatives.

To our knowledge, this is one of the first applied studies in computational social science to systematically evaluate whether open source LLMs fine tuned in one or two languages can generalize to classify both the topic and stance of text written in 13 languages, including several languages that are unseen during fine tuning for each model, under realistic computational and budget constraints. It offers direct empirical evidence that LLMs can learn to perform specific tasks effectively across languages, without relying on familiar linguistic patterns seen during training. This opens the door to efficient multilingual classification without the need to fine tune a separate model for each language. This opportunity is especially relevant given that fine tuning across many languages typically requires substantial computational resources, annotated data and technical capacity that are not always available. Our study offers concrete evidence for how socially relevant tasks, such as monitoring the online debate on immigration on social media, can be achieved through scalable, cross lingual AI systems that bridge LLMs with real world social science research.

Our study makes two significant contributions. First, we contribute novel and robust empirical evidence of the ability of a small decoder only LLM to identify the semantic intent of messages in multiple languages that it has not been fine tuned on. We show that LLMs can accurately classify migration related content across multiple languages following exposure to text in one or two languages via fine tuning. We present evidence of the generalizable potential of LLMs through deploying learned knowledge from a single language to a range of multiple languages at two classification tasks: (i) identifying migration related content; and (ii) identifying anti and pro migration stances in the text.

Second, we make a methodological contribution. We develop a cost efficient and scalable framework for the deployment of open source LLMs for large scale text classification tasks [28]. We demonstrate this approach by fine tuning a lightweight, open source large language model (LLaMA 3.2 3B) using Low Rank Adaptation (LoRA) [29], an efficient method that significantly reduces the number of trainable parameters. The fine tuned model is then quantized to 4 bit precision using the GPT Generated Unified Format (GGUF), which enables efficient deployment on resource constrained hardware. This quantization step reduces the overall model size by about 50 percent compared to the original model, while preserving high classification accuracy and substantially lowering computational requirements. It efficiently optimizes LLMs, resulting in low requirements for computing infrastructure and energy, and in principle enabling LLM deployment even on small CPU based consumer machines. As such, our proposed approach makes LLM deployment more accessible and environmentally sustainable. Such gains represent important evidence for the effective future use of LLMs and address concerns about the high environmental cost, elevated energy consumption and unequal access to LLMs. Taken together, these contributions have the potential to augment the use of LLMs in the social sciences by fine tuning models in a limited number of languages and deploying them to execute analysis across multiple languages and geographies. Such advances can fill gaps in our understanding of human and social behaviors across different cultural and linguistic settings.

# 2 Results

This section presents the results of our evaluation task: assessing whether fine-tuning a lightweight open-source model, LLaMA 3.2 3B, on one or more languages enables it to learn the concept of immigration-related content in a way that generalizes across multiple languages. Specifically, we test the model's ability to identify whether a tweet is about immigration and, if so, determine the stance it expresses ("pro-immigration", "anti-immigration", or "neutral") across both seen and unseen languages.

Figure 1 provides an overview of the workflow used in this study, which consists of six main steps: collection of a sample of tweets, human annotation, LLM fine-tuning and subsequent quantization, tweet classification (both topic and stance), and several logistic regression models to evaluate the classification results across multiple factors.

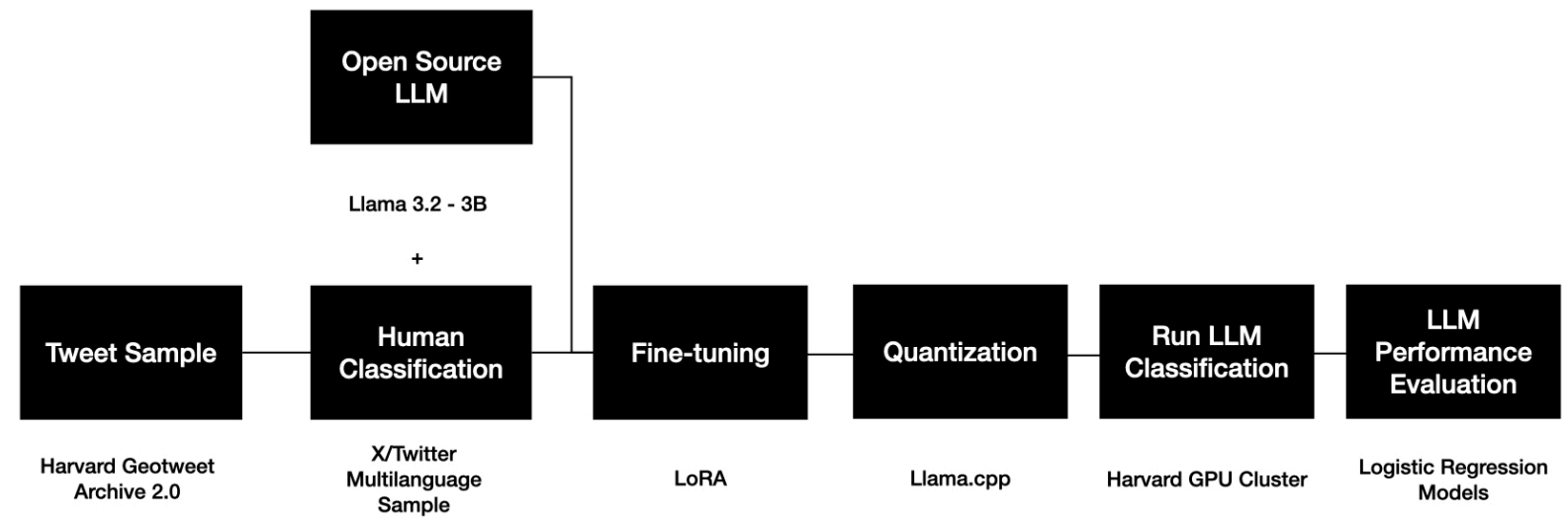

Figure 1: Overview of the workflow used in this study.

Firstly, we extract a sample of tweets in 13 languages from the Harvard Geotweet Archive 2.0, a collection of 10 billion geolocated tweets. These tweets are selected using immigration-related search terms (see Table **??** for details on the tweet search terms used). Each tweet is subsequently labeled by human annotators. Annotators first determined whether a tweet was related or unrelated to immigration. If deemed relevant, the tweet was further classified as expressing a positive (pro-immigration), negative (anti-immigration), or neutral stance towards immigration. Annotation guidelines were iteratively refined to ensure consistency across languages and annotators. Non-English tweets were translated using a LLaMA 3.1 Instruct - 3B quantized 4bit model. Table 3 in the SI Appendix reports the number of tweets annotated per language, showing the linguistic diversity and balance across data used for model training and evaluation.

We developed four fine-tuned LLMs, each trained on a human-annotated data: (1) a monolingual English model, (2) a monolingual Spanish model, (3) a bilingual English–Spanish model, and (4) a multilingual model trained on twelve languages. These four models allow us to compare how fine-tuning on different languages affects generalization to unseen languages and classification accuracy. The monolingual English model is trained using only the English labeled tweets, similarly the Spanish model is trained on Spanish labeled tweets. The English–Spanish model is trained on the combined English and Spanish datasets used in the respective monolingual models. The Multilingual model is trained on this same English and Spanish data, along with additional samples in Arabic, French, German, Hindi, Hungarian, Indonesian, Italian, Polish, Portuguese, and Turkish. The Korean sample is not included in the training of any of the four LLMs.

We estimated four logistic regression models to analyze data from all four fine-tuned LLMs and assess the factors associated with the probability that a fine-tuned language model correctly classifies the standing towards of immigration of the tweets. The dependent variable is binary, indicating whether the model's predicted label matches a human-annotated ground truth. In all model specifications, the baseline corresponds to a model fine tuned on English language tweets, applied to untranslated input, with the tweet labeled as Neutral in terms of immigration stance by the human annotators, written in English, and belonging to the training set. The regression intercept reflects the classification accuracy under this baseline condition. All other coefficient estimates represent deviations from this reference point as log-odds.

Figure 2 presents the results from a baseline logistic regression model (Model 1) predicting the likelihood of correct tweet classification, expressed in log-odds, as a function of model type, label category, input language, translation quality, and whether the tweet is part of the test or training dataset split. The full coefficient estimates are reported in Table 7 in the SI Appendix.

Tweets labeled as `Unrelated` are significantly easier to classify compared to the baseline category `Neutral` ($\beta = 1.579$, $p < 2 \times 10^{-16}$), followed by tweets expressing pro-immigration ($\beta = 0.222$, $p < 3.7 \times 10^{-8}$) and anti-immigration views ($\beta = 0.215$, $p < 1.1 \times 10^{-7}$). There is notable variation across languages. Tweets written in Polish ($\beta = -1.034$, $p < 2 \times 10^{-16}$), Hungarian ($\beta = -0.865$, $p < 2 \times 10^{-16}$), and Italian ($\beta = -0.710$, $p < 2 \times 10^{-16}$) are substantially harder to classify than English tweets, which serve as the reference category.

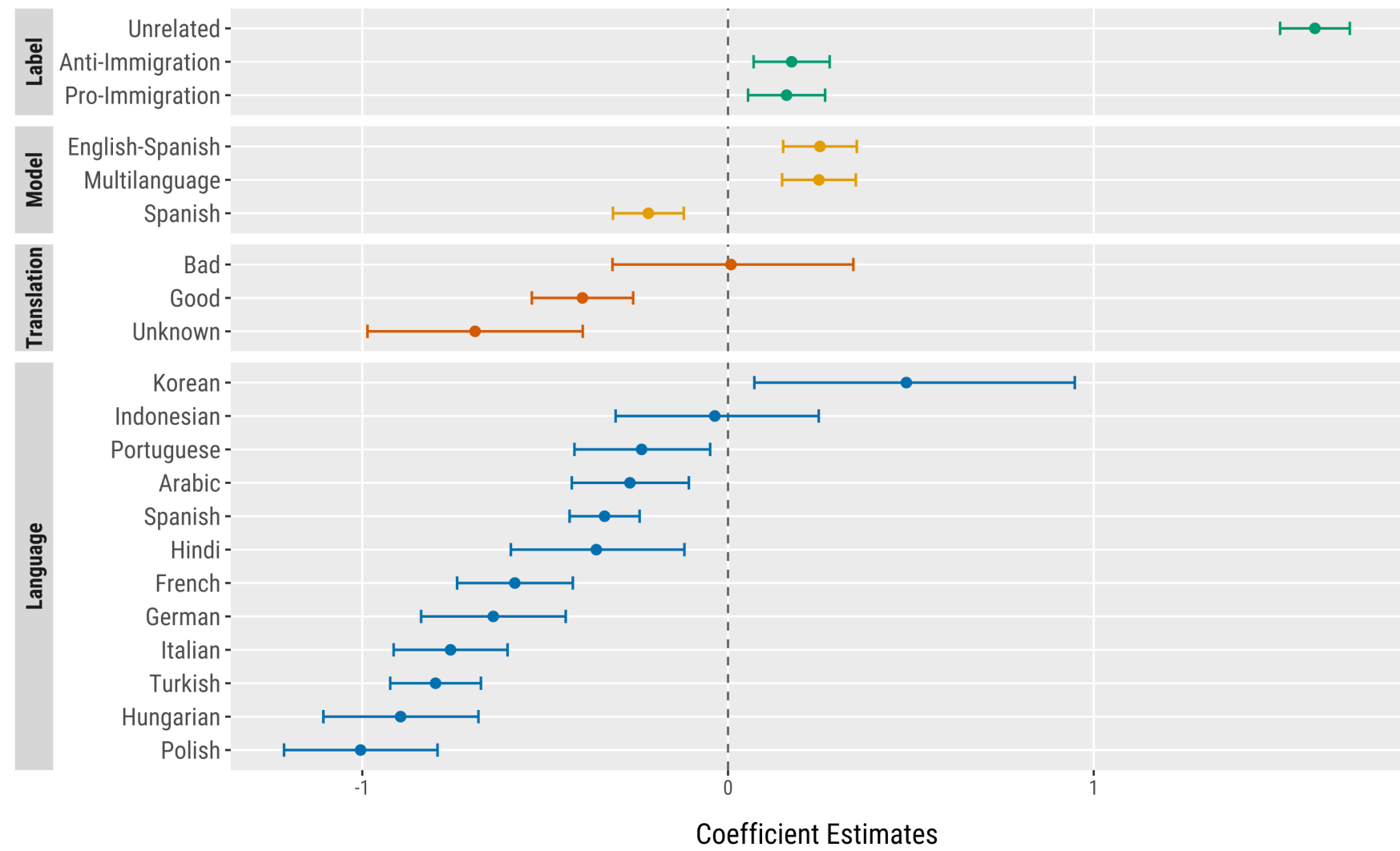

Figure 2: **Estimated coefficients (log-odds) from the baseline logistic regression model.** The figure displays point estimates and 95% confidence intervals for the main effects of *Model* type, tweet *Language*, *Translation* quality, and *Label.* The model estimated the probability that a fine-tuned language model correctly classifies a tweet's stance toward immigration. Coefficients are expressed in log-odds and grouped by covariate type. Our baseline model is the one fine-tuned exclusively on English data, evaluated on untranslated tweets labeled "Neutral" from the "Train" set. Estimates are derived from Model 1, which includes no interaction terms (Eq. 3). The log-odds coefficient for *Train/Test* is excluded from the figure.

The bilingual English-Spanish model ($\beta = 0.280$, $p < 3.3 \times 10^{-13}$) and the Multilingual model ($\beta = 0.263$, $p < 6.4 \times 10^{-12}$) both outperform the English-only baseline, highlighting the benefits of multilingual exposure during fine-tuning as well as a larger size of training dataset. On the other hand, the Spanish-only model performs significantly worse ($\beta = -0.215$, $p < 5 \times 10^{-8}$) compared to the English model baseline.

Translation significantly reduces classification accuracy, regardless of assessed quality. Tweets that were machine-translated from other languages into English tend to be harder to classify correctly than their original, untranslated counterparts. Even tweets marked as `Good` translations show a significant decrease in the ease of classification ($\beta = -0.381$, $p < 3.9 \times 10^{-8}$), while `Unknown` translations perform worse still ($\beta = -0.680$, $p < 1.2 \times 10^{-6}$). Translations classified as `Bad` do not show a statistically significant effect, likely due to limited sample size. Translation likely introduces semantic misalignments as seen in previous research [16].

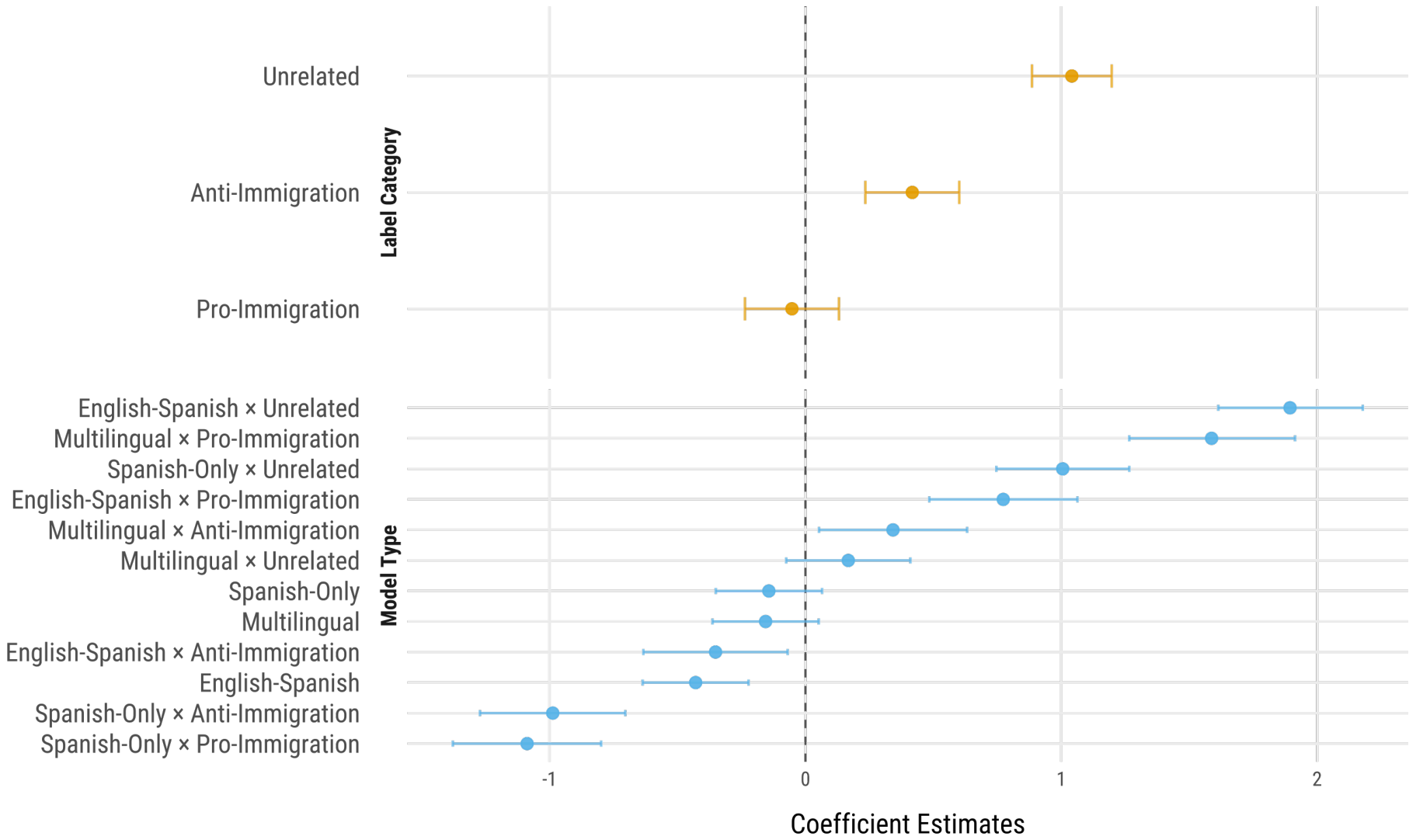

Figure 3: **Estimated coefficients (log-odds) of model type and label category on classification accuracy.** The figure shows coefficient estimates for *Model* and *Label* main effects and their interaction terms from Model 2 (Eq. 4). The model estimated the probability that a fine-tuned language model correctly classifies a tweet's stance toward immigration. The full model also includes controls for *Language*, *Translation Quality*, and *Train/Test* split, which are not displayed in this figure. Estimates are presented in log-odds, with horizontal lines denoting 95% confidence intervals. The baseline category represents the English model applied to untranslated English tweets labeled as "neutral" in the training set, with all coefficients showing effects relative to this reference point.

### Fine-tuning teaches the topic regardless of language

We investigated whether the impact of fine-tuning varies not only by model type but also by the type of label, specifically, whether LLMs differ in their classification accuracy for `unrelated`, `anti-immigration`, and `pro-immigration` human-annotated tweets. To test this, we developed a new model (Model 2), which extends Model 1 by including interaction terms between `Model` and `Label` (Eq. 4).

Figure 3 visualizes the estimated coefficients from Model 2, highlighting how each model performs across different label categories (see Table 8 in the SI Appendix for the full set of regression coefficients).

The results show a clear distinction in model strengths. The English-Spanish model performs significantly better ($\beta = 1.755$, $p < 2 \times 10^{-16}$) than the others at classifying unrelated content. This suggests that, even without being fine-tuned on multiple languages, the model is able to learn the general topic of immigration and distinguish unrelated tweets from immigration-related ones even in languages it was not exposed to during training.

In contrast, the Multilanguage model excels in detecting specific stances. It shows a strong positive effect in classifying both anti-immigration content ($\beta = 0.198$, $p = 0.084$) and especially pro-immigration content ($\beta = 1.622$, $p < 2{\times}10^{-16}$). These coefficients indicate that fine-tuning on multiple languages supports better sensitivity to subtle distinctions in sentiment or stance, even if overall topic understanding transfers across language boundaries.

Taken together, these results show that fine-tuning enables LLMs to learn the general topic of immigration regardless of the languages used during training. This is evident in the English-Spanish model's strong performance in identifying unrelated content, even in languages it was not explicitly trained on. However, accurately detecting specific attitudes toward immigration, particularly pro-immigration stances, benefits from multilingual exposure, likely because the expression of ideological positions varies across languages.

These findings support the broader claim that while understanding whether content is about immigration is largely transferable across languages, capturing the nuances of ideological stance in multilingual data requires exposure to linguistic diversity during fine-tuning. This improvement reflects the combined effect of multilingual exposure and the increased training sample size that accompanies it, aligning with fundamental statistical principles that larger, well-curated training sets yield better model performance.

### Accuracy of fine-tuned models varies by language

To investigate how classification performance varies across languages and model types, we estimated a logistic regression model (Model 3) with main effects for `Model` and `Language`, along with their interaction. This allows us to assess how each model performs in different linguistic settings while controlling for `Label`, `Translation` and `Train/Test` similarly to Model 1.

Figure 4 displays a heatmap of log-odds for each combination of model and language based on estimates from Model 3. The visualization highlights substantial variation in model performance across linguistic contexts, derived from the main and interaction effects between fine-tuned model type and language. The full set of coefficient estimates from Model 3 is reported in the SI Appendix (Table 9).

We calculated total log-odds by summing the coefficients for each model, language, and their interaction. We report the cumulative log-odds for each language and model in Table 9 in the SI Appendix.

The Multilanguage model performed better than other models in most languages, especially in Polish (–0.718) and Turkish (–0.573). This suggests that multilingual fine-tuning enabled the model to generalize better in challenging settings, even when classification difficulty remained high in absolute terms.

The English-Spanish model performed best in Indonesian (0.435) and Korean (0.545), the only two languages where any model achieved a positive total log-odds score. These results indicate that even without explicit exposure to these languages, the model successfully transferred topic understanding, likely due to structural or topical similarities in the training data.

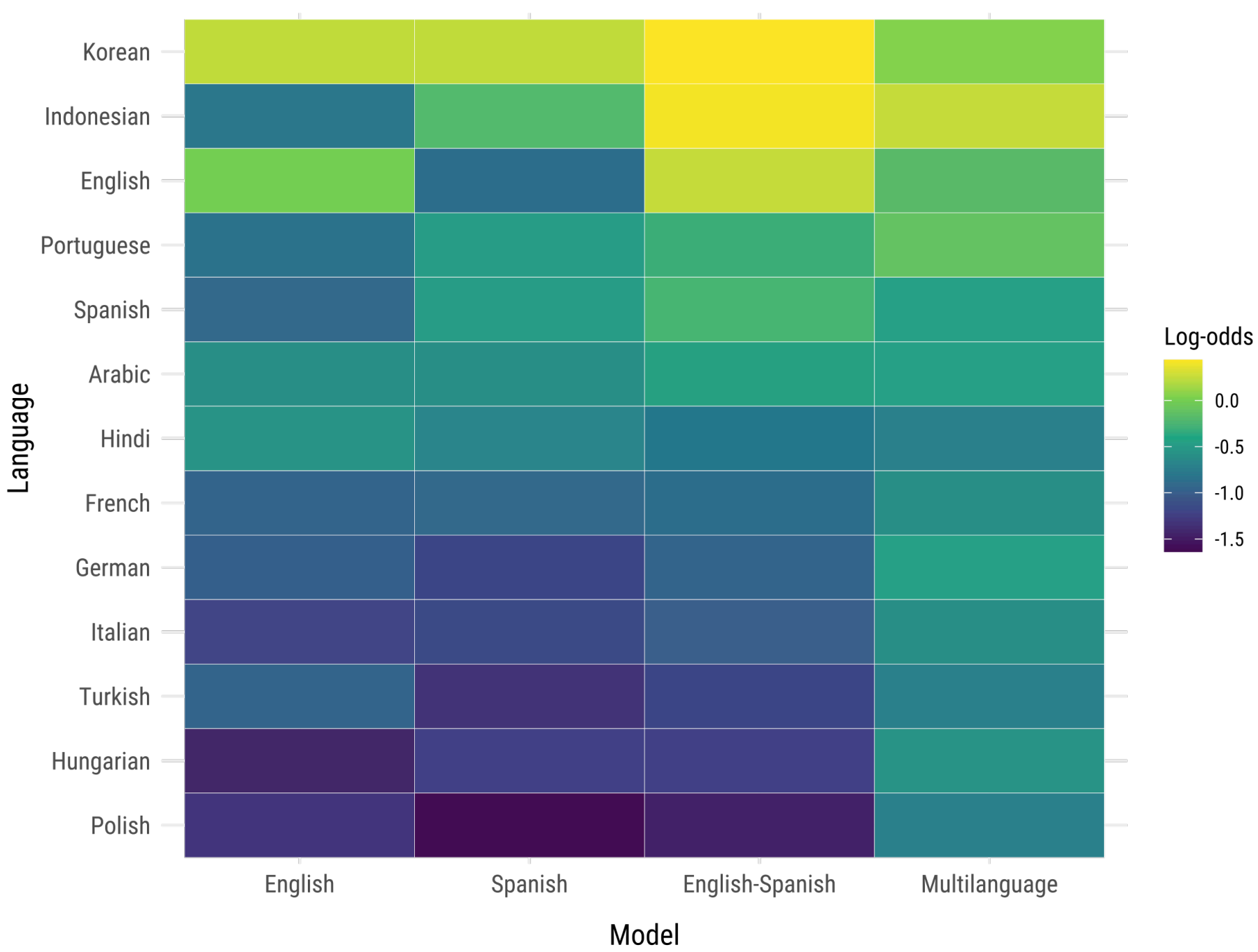

Figure 4: **Cumulative log-odds of correct classification by model and language** Heatmap shows predicted classification accuracy for each combination of language on the y-axis and model on the x-axis (English, English–Spanish bilingual, Multilanguage, and Spanish), based on coefficient estimates from Model 3 (Eq. 5). For each cell, the total effect is computed as the sum of main and interaction terms involving the selected model and language, with all other covariates held constant. The darker the cell, the lower the predicted log-odds of correct classification for that model–language pair.

The Spanish-only model underperformed across the board. Interestingly, even in Spanish, its total effect (–0.468) was more negative than that of the English-Spanish model (–0.272). This suggests that the English-Spanish model benefited from a larger training dataset and a more focused multilingual exposure, making it better suited than the Spanish-only model, even for classifying tweets in Spanish. We report the complete F1 scores in Table 5 in the SI Appendix.

These findings suggested that while the Multilanguage model generally achieved higher classification performance in languages seen during fine-tuning, the English-Spanish model performed particularly well in languages like Indonesian and Korean, suggesting that topic knowledge could transfer effectively across languages, even without explicit exposure.

### Models fine-tuned in fewer languages are more accurate in identifying topic-relevant content

Label distributions varied notably across languages. For example, Korean tweets contained a substantially higher proportion of unrelated content than others which may complicate the interpretation of language-specific accuracy estimates. While Model 2 incorporated interactions between model and label, it did not address how differences in label distribution across languages might affect classification outcomes. In particular, previous models were unable to disentangle whether high performance in certain languages reflected true model understanding or simply an easier underlying label mix. Both Models 1 and 2 showed that unrelated tweets were generally easier to classify, and Model 3 revealed variation in performance across language–model combinations. This raises an important issue: when a model performs well in a given language, is it due to its ability to interpret that language's immigration discourse, or because the dataset contains a disproportionately high number of unrelated tweets? For example, the strong performance of the English–Spanish model on Korean tweets may not indicate a deep understanding of Korean-language immigration content, but rather reflect the overrepresentation of unrelated tweets in that subset.

To address this limitation, we introduce a variable called `ShareUnrelated`, which captures the proportion of tweets labeled as unrelated to immigration within each language subset of our dataset. For example, if 70% of Korean tweets are classified by human annotators as `unrelated`, then every Korean-language observation is assigned `ShareUnrelated` = 0.70. This variable allows us to control for variation in label composition across languages and assess whether model performance is influenced by the relative ease of classifying unrelated content.

We estimated a fourth logistic regression model (Model 4) that included the interaction term between `ShareUnrelated` and `Model`. This model builds on Model 3 by retaining the interaction between `Model` and `Language`, as well as `Translation` and `Train/Test` as control variables. By incorporating these additional factors, Model 4 allows for a more nuanced understanding of how both model architecture and dataset composition influence classification performance across languages. Table 10 reports the complete log-odds from Model 4.

Figure 5 shows the compounded effect by summing three components: the `Model`'s main coefficient, the main effect of `ShareUnrelated`, and the interaction term between `Model` and `ShareUnrelated`. This total reflects the predicted performance (in log-odds) when the input data is dominated by tweets not related to immigration.

The English-Spanish model produced the highest compounded effect, with a total of 5.009, and all its contributing coefficients were statistically significant. The Spanish model followed with a total effect of 4.456, although its main effect was not statistically significant. The Multilanguage model had the lowest total effect at 3.597, and neither its model coefficient nor its interaction term was statistically significant.

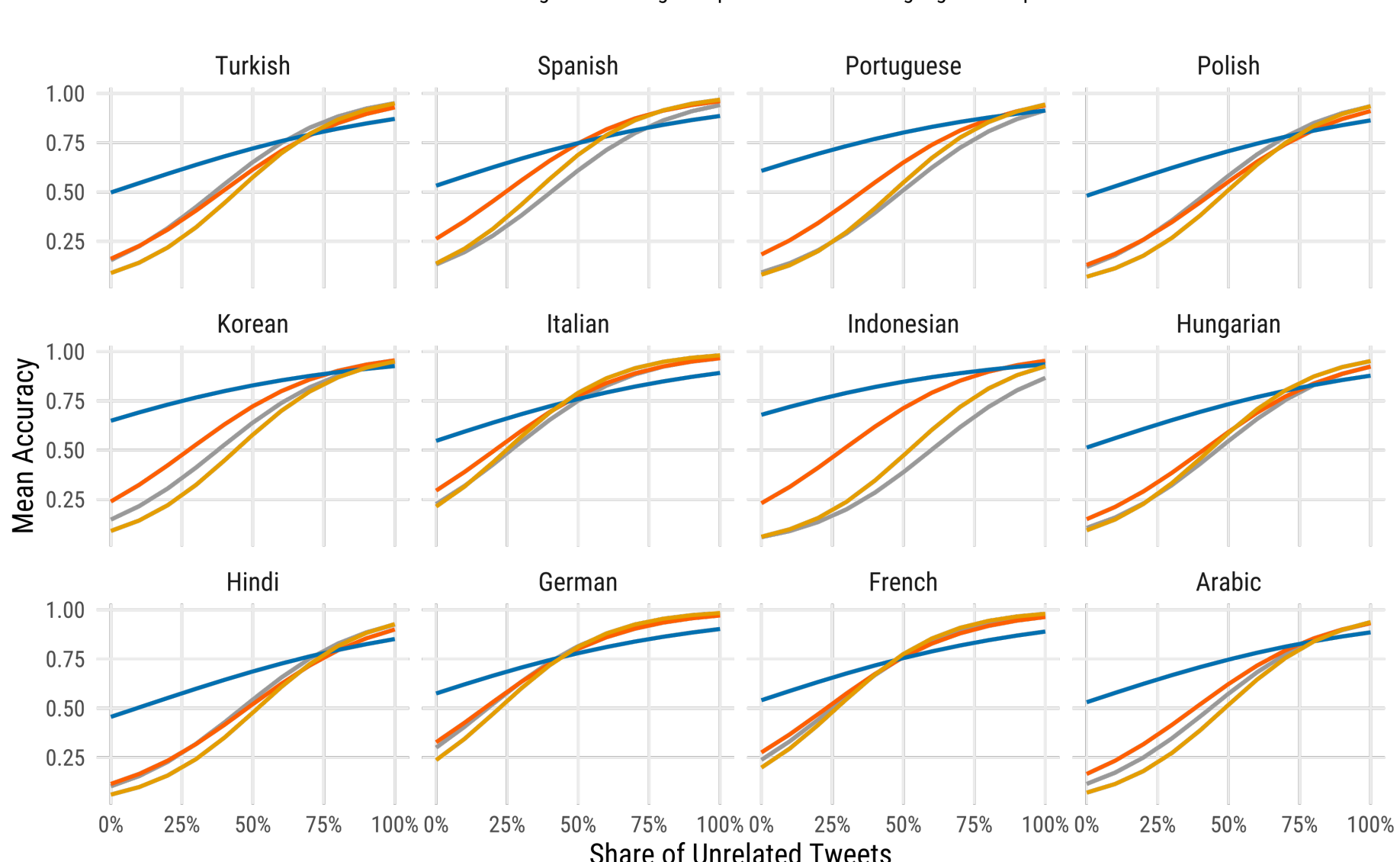

Figure 5: **Predicted classification accuracy of fine-tuned language models as the share of unrelated tweets ranges from 0% to 100%.** Accuracy curves are generated by inserting the estimated coefficients from the interaction model (Eq. 6) into the logistic link function $\text{Accuracy} = 1/\left(1+e^{-\,\textit{Total Effect}}\right)$, where *Total Effect* combines main and interaction terms for *Model*, *Language*, and *Share Unrelated*. Lines correspond to four model variants (English, English–Spanish, Multilanguage, and Spanish) and panels separate the language of the tweets being classified.

These findings suggest that in contexts where a large share of tweets are unrelated to immigration, the English and Spanish models tend to outperform the multilingual model. This may reflect a trade-off: while multilingual fine-tuning improves a model's ability to detect ideological stance within immigration-related content, it may reduce its robustness when dealing with off-topic data.

More broadly, the results show that models fine-tuned in only one or two languages can still learn to identify whether a tweet is about immigration. This suggests that the ability to recognize topic relevance can generalize across languages. In contrast, the multilingual model performs best when tweets are clearly about immigration, as it is more effective at distinguishing between nuanced stances. However, this strength may come at the cost of lower performance when processing unrelated or ambiguous content, where the English and Spanish models demonstrate greater reliability.

### Fine-Tuning Reduces LLM Pretraining Biases

An important determinant of a fine-tuned LLM's multilingual classification performance is the extent to which each language was represented in its original

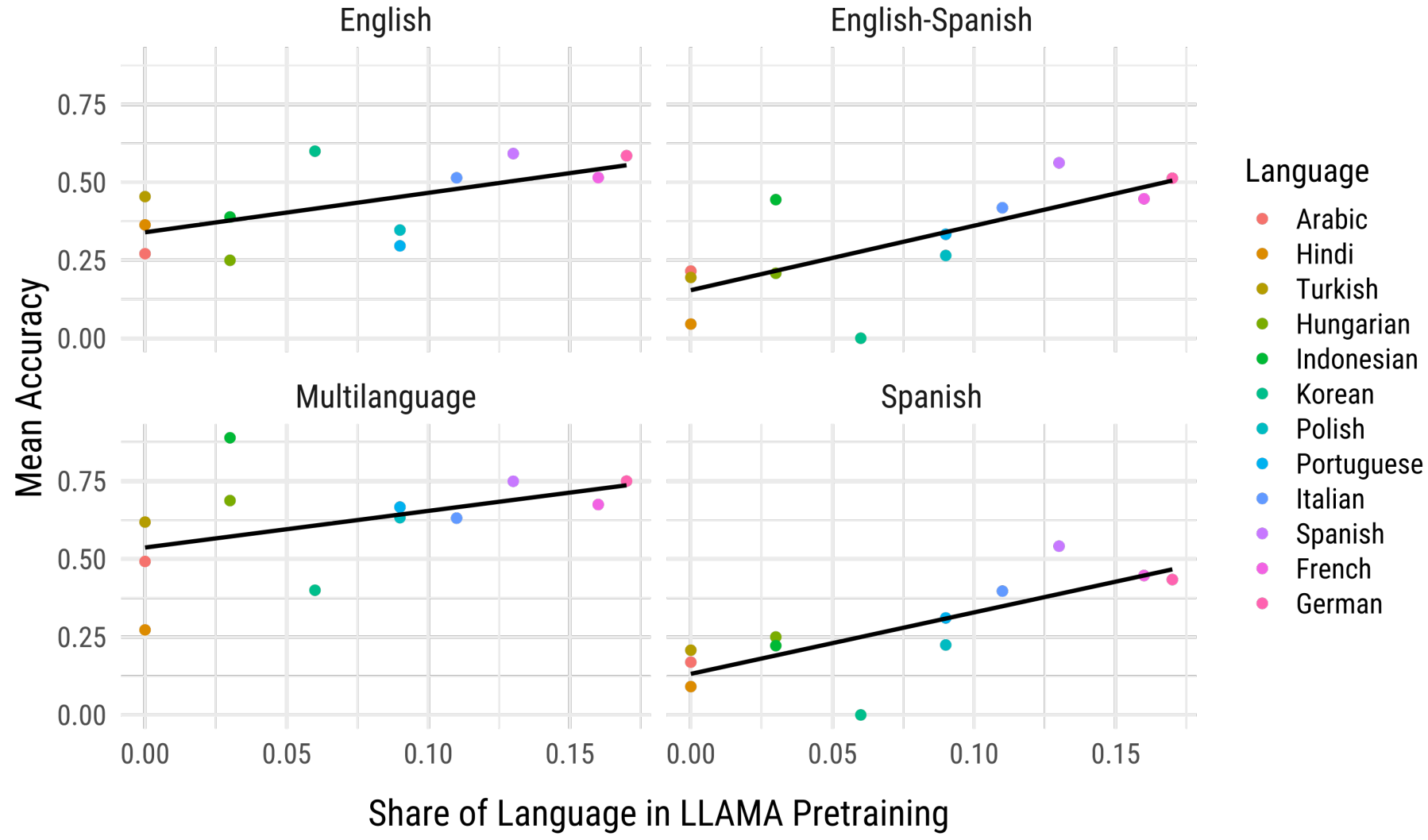

Figure 6: **Classification accuracy increases with pretraining language exposure across all model variants.** Each point represents mean accuracy for a language-model combination, plotted against the share of that language in the pretraining corpus. The black regression line shows the overall positive relationship between pretraining exposure and classification performance. Languages in the legend are ordered by the lowest share (eg. Arabic) in the LLaMA corpus to the highest (eg. German)

pretraining corpus. Pretraining data are known to shape a model's general language knowledge [30],[31], and heavily imbalanced exposure can lead to systematic performance gaps across languages [32]. Since LLMs were trained predominantly on English texts [33], we hypothesized that languages with greater presence during pretraining are likely to show higher classification accuracy. To examine this structural dimension, we rely on language share estimates from the LLaMA 2 pretraining corpus, as Meta has not publicly released detailed language composition data for the LLaMA 3 family of models which includes LLaMA 3.2 - 3B, the lightweight model that we have used for fine-tuning. While the exact figures for LLaMA 3 remain unknown, Meta has reported that only 8% of its pretraining tokens are non-English, closely aligning with the 11% reported for LLaMA 2. Given this similarity, we used the LLaMA 2 reported languages as a reasonable proxy to infer the likely distribution of language representation in LLaMA 3 models and to assess the potential impact of pretraining biases on model performance across languages. We assigned to each language in our dataset, the share of that language that is present in the pretraining corpus (see Table 4 in the SI Appendix for details).

Figure 6 illustrates how classification accuracy varies by model type, language, and the percentage of language (except English) present in the pretrain-ing corpus, as reported by Meta. To calculate the accuracy shown in the figure, we filtered out English tweets and tweets labeled as “unrelated“, grouped the data by language, model type, and pretraining language share, and computed the mean accuracy for each group based on the correct observations. The figure reveals that, for all LLMs, accuracy tends to be higher for languages with greater pretraining exposure. This pattern is particularly strong for the monolingual and bilingual models, suggesting that pretraining exposure influences accuracy across languages. This points to an underlying and potentially problematic bias in the text classification task based on the corpus of texts used for LLaMA’s pretraining.

However, our fine-tuned Multilanguage LLM displays a more stable level of classification accuracy even for languages with limited or no representation at the pretraining stage, such as Hungarian and Turkish. This may be explained by the Multilanguage LLM exposure during the fine-tuning phase to these under-represented languages which reduces the difference in accuracy generated by the language bias in the pretraining. These results suggested that introducing as few as 3000 tokens out of 15.6 trillion tokens of the pretraining, can meaningfully boost classification accuracy for that language and help counter the English-centric bias of today’s open- and closed-source LLMs.

### Fine-tuned 3B model processes tweets 26–168× faster and 109–8,690× cheaper than commercial LLM APIs such as ChatGPT or DeepSeek

Beyond classification accuracy, our approach offers substantial advantages in terms of computational efficiency, cost, and environmental impact. Table 6 presents a comprehensive comparison of resource requirements for classifying 10 billion tweets using our fine-tuned Llama 3.2-3B model versus leading proprietary alternatives. Our fine-tuned model, running on Harvard’s FASRC GPU cluster, achieves remarkable efficiency gains across multiple dimensions. Operating at 3,854 tokens per second, it processes text between 24 to 168 times faster than proprietary alternatives, with the speed advantage being most pronounced compared to DeepSeek models (143× faster than DeepSeek Chat, 168× faster than DeepSeek Reasoner) and substantial even against the fastest proprietary models like GPT-4.1 (26× faster). The cost implications are even more dramatic. We estimate that classifying the entire Harvard 10 billion Geotweet Archive 2.0 dataset would cost just $2,900 using our approach, compared to $316,000 for the least expensive proprietary alternative (DeepSeek Chat) and up to $25.2 million for OpenAI GPT 5.2 Pro. This represents cost reductions of 109× to 8,690× compared to commercial APIs. Environmental benefits are equally significant. Our model, powered by Harvard’s renewable energy infrastructure, produces virtually zero carbon emissions, while proprietary alternatives generate between 271 to 3,139 tonnes of $CO_2$ for the same classification task. Water usage follows a similar pattern, with our approach requiring only 14.3 $m^3$ compared to 347.7 to 37,750.4 $m^3$ for commercial models. These efficiency gains stem from several technical optimizations: (1) the use of a smaller, more efficient 3B parameter model compared to much larger proprietary alter-

natives (175B to 1.7T parameters), (2) 4-bit quantization that reduces model size by 50% while preserving accuracy, (3) LoRA fine-tuning that minimizes trainable parameters, and (4) deployment on renewable energy infrastructure. Importantly, these optimizations do not compromise classification performance, as demonstrated by our accuracy results across languages and tasks. The resource efficiency of our approach addresses a critical barrier to multilingual LLM research: the assumption that high-quality performance requires either expensive proprietary models or computationally intensive training of separate models for each language. Our results demonstrate that a single, efficiently fine-tuned open-source model can achieve competitive multilingual performance while dramatically reducing financial, computational, and environmental costs.

# 3 Discussion

A central finding of this study is that fine-tuning large language models enables them to internalize topic understanding in a way that generalizes across languages and cultural contexts. Even when trained on data in just one or two languages, LLMs can learn to recognize whether content is topically relevant, in this case, whether a tweet is about immigration, including in languages the model has never seen during fine-tunining training. This suggests that topic understanding is not tied to specific linguistic patterns but reflects deeper semantic understanding that can transfer across multilingual environments. For example, our bilingual English–Spanish model performed well in identifying unrelated content across a range of typologically diverse languages, reinforcing the idea that topic comprehension is broadly transferable.

However, when the task shifts from recognizing topic relevance to classifying ideological stance, such as distinguishing between pro- and anti-immigration tweets. performance becomes more variable. The multilingual model, fine-tuned on a diverse set of twelve languages, consistently outperformed the monolingual and bilingual models in this more nuanced task, particularly in low-resource or typologically distant languages. Yet this improved sensitivity to ideological variation comes with a trade-off: the multilingual model is less accurate when classifying datasets with a high proportion of unrelated content. In such cases, simpler models trained on fewer languages, like the English–Spanish model, are more robust in filtering relevant from irrelevant content, likely because their training data offer a more focused signal. By accounting for the share of unrelated content in each language dataset, we observed that monolingual and bilingual models often outperformed the multilingual model when it came to identifying whether tweets were actually about immigration. This suggests a key trade-off: the multilingual model is more effective at detecting subtle differences in stance toward migration but less reliable at distinguishing relevant from irrelevant content, while simpler models are better at filtering off-topic tweets but less accurate in identifying ideological positions.

A further key insight is that imbalances in LLaMA's pre-training language distribution influences its downstream performance. Languages with greater presence in the LLaMA pretraining corpus, such as German, French, and Spanish, tended to show higher classification accuracy across all models, particularly the monolingual and bilingual ones. This aligns with previous work suggesting that foundation model behavior is shaped not just by fine-tuning, but also by the structure and balance of the original pretraining data. However, encouragingly, we also find that these biases can be meaningfully mitigated. The Multilingual model achieved relatively stable performance even in languages like Hungarian and Turkish, which are underrepresented or absent from the pretraining corpus. In several cases, exposing the model to as few as 75 annotated tweets — equivalent to $8.65 \times 10^{-11}$ of the 15.6 trillion tokens used to pretrain LLaMA 3 — was enough to yield clear gains in classification accuracy. This demonstrates the disproportionate value of targeted fine-tuning, especially for low-resource languages.

Finally, our analysis highlights the cost of relying on translation as a strategy for multilingual classification. Tweets translated into English, regardless of the assessed translation quality, were classified less accurately than their original counterparts. Even "good" translations led to degraded model performance, suggesting that critical semantic or stylistic cues are lost in translation. This effect persisted even when using high-quality human-rated translations. These findings reinforce the importance of training models directly on native-language content rather than translating it into English. Exposure to the original linguistic and cultural framing appears necessary for models to fully capture the meaning of online discourse, particularly when it involves ideologically sensitive topics like immigration.

Together, these findings underscore both the promise and limits of multilingual generalization in fine-tuned LLMs. Topic-level understanding appears transferable even with minimal language exposure. However, for tasks requiring ideological nuance, multilingual fine-tuning is essential. It not only improves performance on seen languages but also enhances generalization to underrepresented and typologically distant ones. Crucially, this can be achieved even with very small, curated datasets, suggesting a practical path forward for improving equity in multilingual NLP and computational social science research.

## 4 Conclusions

The rapid adoption of LLMs is transforming how social scientists analyze text across languages, scales, and social domains. These tools promise efficiency and reach. Yet our understanding of how they function in multilingual, real-world settings remains limited, particularly in complex and culturally embedded issues like immigration. Much of the existing research relies on English-centric approaches, synthetic data, and machine translation strategies that can obscure meaning and cultural nuance. Furthermore, existing studies often evaluate performance using standardized benchmarks that may not reflect the linguistic complexity and cultural specificity of real-world social science applications. Additionally, the field lacks systematic comparisons of monolingual versus multi-lingual fine-tuning approaches for culturally sensitive topics where linguistic and cultural context critically influence meaning. While fine-tuning separate monolingual models for each language appears to be one viable strategy for achieving optimal performance, this approach presents significant practical barriers for most social science research contexts: it requires multiple human-annotated datasets, substantial computational resources for training multiple models, and considerably longer development timelines. These resource constraints may inadvertently limit multilingual research to well-funded institutions or force researchers to accept suboptimal performance across languages. The literature also provides limited guidance on how performance disparities across languages might compound existing inequalities in whose voices and perspectives are accurately captured in computational social science research. Moreover, while recent work has documented language representation imbalances in pretraining corpora, there has been limited empirical investigation of how these structural biases interact with fine-tuning strategies to shape performance across languages in domain-specific classification tasks. These knowledge gaps has significant implications for the design, deployment and ethical use of LLMs in the social sciences.

Our study addresses these gaps by examining how fine-tuning strategies affect multilingual classification performance in immigration discourse. We found that fine-tuned LLMs can successfully identify immigration-related content even in languages not seen during training, suggesting that topic-level knowledge generalizes across languages through the model's deeper semantic representations. However, more nuanced tasks like stance detection—distinguishing supportive, opposing, or neutral positions—require richer linguistic exposure. These tasks rely on language-specific surface features such as morphology, orthography (eg. writing in capital letters), and cultural idioms (eg. "build the wall") that vary significantly across languages. Consequently, multilingual fine-tuning substantially improves stance classification accuracy, particularly for underrepresented or typologically distant languages.

We also demonstrated that pretraining corpus composition creates systematic performance biases. Languages with greater representation in the original training data consistently achieved higher classification accuracy, though these gaps can be substantially reduced through targeted multilingual fine-tuning. We also found that the fine-tuned model labels tweets more accurately in their original language than in high-quality translations, indicating that translation introduces semantic noise.

We acknowledge some limitations. Our study focuses on immigration discourse and short-form social media content. Future research should examine whether these patterns hold across different domains and text types. Additionally, while we identified associations between pretraining exposure and performance, further investigation is needed to establish causality and explore how cultural and discursive dimensions interact with fine-tuning strategies. Despite these limitations, our work provides actionable insights for building more inclusive and practical multilingual classification systems in computational social science.

Our findings offer both methodological and practical contributions. Methodologically, our work demonstrated that domain-specific knowledge can generalize beyond the training language, suggesting a scalable path forward for multilingual LLM applications without exhaustive data requirements. Practically, we showed that fine-tuning open-source LLMs using resource-efficient techniques like LoRA and 4-bit quantization offers a remarkably cost-effective and environmentally sustainable alternative to proprietary systems. Our fine-tuned Llama 3.2-3B model, running on Harvard's FASRC GPU cluster, operates at 3,854 tokens per second—between 36 to 168 times faster than proprietary alternatives. We estimate that our approach could classify the entire Harvard 10 billion Geotweet Archive 2.0 dataset for just $2.9k with virtually zero carbon emissions, compared to proprietary alternatives that would cost between $316,000 and $17.7M while generating hundreds of tonnes of $CO_2$. This lightweight framework enables scalable multilingual classification and dramatically lowers barriers to adoption for research teams with limited computational budgets, supporting broader efforts to democratize AI.

# 5 Materials and Methods

This study explores how fine-tuning a lightweight open-source LLM on human-annotated data improves its ability to classify tweets about immigration across multiple languages. Our approach combines three main components: the collection and annotation of multilingual tweets, the fine-tuning of models with different levels of language exposure, and a structured evaluation of classification performance across topic and stance. Immigration serves as an ideal test case due to its global relevance, ideological complexity, and prominence in online discourse, where it often triggers polarized debate and targeted hostility. By using human annotations to validate model predictions, we assess whether fine-tuning enables LLMs to learn the concept of immigration in a way that generalizes across linguistic and cultural boundaries.

## Open Source LLM: LLaMA 3.2, 3B Parameters

We built this model to classify immigration-related content in multilingual tweets, with the goal of deploying it on the Harvard Geotweet Archive 2.0, which contains over 10 billion tweets. To our knowledge, this is by far the largest publicly available social media dataset, more than 5 times the size of the 2 billion tweet dataset developed by Imran et al. [34]. This scale required a model that is both efficient and scalable. We fine-tuned the LLaMA 3.2 model with 3 billion parameters, an open-source model released by Meta AI in 2024. We selected LLaMA over commercial alternatives such as ChatGPT (OpenAI) or Claude (Anthropic) for several reasons. First, open-source models like LLaMA offer full access to model weights and architecture, enabling fine-tuning and modification at all levels, an essential feature for replicable research and understanding how models learn. In contrast, proprietary models limit fine-tuning to constrained API-based methods and do not support retraining of core model layers. Second, using closed APIs at this scale would have been prohibitively expensive, with inference alone estimated between $316,000 and over $17 million (see Table 6 in the SI Appendix). By combining an open-source model with university cluster infrastructure, we dramatically reduced the cost of training and inference. Third, using an open-source model allowed us to release the resulting classifier publicly, supporting open science and enabling reuse in other domains. Finally, we optimized the model for deployment by converting it into GGUF format and applying 4-bit quantization, which reduced memory usage and inference time with minimal impact on accuracy.

Among the open-source options available at the time of processing (Fall 2025), LLaMA 3.2 offered one of the strongest multilingual foundations, having been trained on 15.6 trillion tokens across multiple language, an order of magnitude more than earlier versions like LLaMA 2, which was trained on only 1.5 trillion tokens. We selected the 3 billion parameter version to balance predictive performance with practical constraints around speed, memory, and deployment cost. As shown in Carammia et al. (2024) [28], small fine-tuned models can perform as well as or better than much larger base models from the same family or across model families. For example, a fine-tuned LLaMA 3 model with 3 billion parameters outperforms a 70 billion parameter base model, while being significantly more efficient in terms of inference speed and memory usage. Moreover, the marginal performance gains from fine-tuning very large models tend to diminish, making smaller models more attractive for large-scale classification. These combined factors made LLaMA 3.2, 3B a robust and scalable solution for multilingual tweet classification at the scale of the Harvard dataset.

The substantial differences in both pretraining scale and model architecture make newer LLMs qualitatively different from BERT, limiting the degree to which insights from earlier work can be extended to LLaMA style models. Notably, LLaMA models already outperform BERT based transformers in multilingual hate speech classification in Hindi-English particularly for implicit or indirect abuse a feature that the online immigration debate is especially prone to [12]. Yet, this has not been tested for classifying content related to immigration across multiple languages and at scale.

Taken together, these findings indicate that while BERT family models remain useful historical baselines, modern decoder-only LLMs such as LLaMA 3 are far more suitable for multilingual topic and stance classification, especially in the context of immigration discourse and at the scale required for operational deployment across trillions of tokens.

## Fine-tuning: LoRa

Fine-tuning allows us to specialize LLMs for specific tasks without retraining from scratch. We fine-tuned the LLaMA models using Low-Rank Adaptation (LoRA), which modifies the model's output to better suit our particular classification task. Unlike proprietary models, such as ChatGPT, open-source fine-tuned models can be freely shared, advancing reproducible (social) science.

For fine-tuning, we used Harvard's Faculty of Arts & Sciences Research Computing (FASRC) high performance computer cluster including A100 and H100 GPUs. For the classification tasks we used A11 and H100 GPUs. Fine-tuning and classification were performed using spare cycles of the cluster (i.e. when GPUs where idling). This required optimized strategies of check-pointing both fine-tuning and inference tasks as our processes could get terminated by legitimate GPU owners at any time during the execution. In this way, our project did not generate additional CO2 compared to the standard usage of an existing cluster.

## Translation

To assess the impact of translation within a multilingual classification pipeline, we employed a 4-bit quantized version of the LLaMA 3.2 3B Instruct model to translate tweets from non-English languages into English. While larger models (eg. LLaMA 3.1 70B) may offer higher translation fidelity, their resource demands make them impractical for large-scale use since they would require extensive GPU infrastructure, which is inconsistent with our lightweight and reproducible inference pipeline. Similarly, commercial LLM APIs would require massive financial resources with an estimated cost around $4.5 million just for translating 10 billion tweets using the Open AI GPT 4o model. As result, our translation approach ensures consistency with the speed, computational and financial constraints of our broader evaluation workflow.

## Datasets

To build the logistic regression models, we constructed the dataset by aggregating classification outcomes from four fine-tuned models: English-only, Spanish-only, English-Spanish bilingual, and a Multilanguage model. Each of these LLM have classified all the 4,900 human-annotated tweets. Additionally, the English-only LLM classified 1100 non-English tweets which were machine-transalted to measure the impact of the translation on the accuracy outcomes (see Table 2 in the SI Appendix).

Each observation in our dataset has the following variables:

- **Tweet language** (e.g., English, Spanish, Arabic),
- **Tweet label** (pro-immigration, anti-immigration, neutral, or unrelated),
- **Translation quality**, based on human assessment (`good`, `bad`, `unknown`, or `not translated`),
- **Model used to classify the tweet**, indicating the language(s) it was fine-tuned on,
- **Dataset split**, denoting whether the observation came from the training or test set.

The binary dependent variable was coded as 1 if the predicted label matched the human-assigned label, and 0 otherwise. Additionally, we have assigned to each tweet `Language`, the reported proportion of that language in the pretraining corpus of LLaMA 2, the foundational open source model used in the fine-tuning. To estimate language coverage in pretraining, we relied on publicly reported statistics for LLaMA 2 since granular data for LLaMA 3 have not been released. Meta reports that approximately 8% of its pretraining tokens are in non-English languages (Meta, 2024). In contrast, LLaMA 2's pretraining corpus contains about 10.3% non-English content. While the specific language distributions may differ between the two models, this comparison provides a reasonable proxy for estimating language exposure. The language distribution from LLaMA 2's pretraining corpus is reported in Table 4 in SI Appendix.

## Model Specifications

To assess model performance across different linguistic and experimental conditions, we fit a series of logistic regression models predicting the probability that a tweet is classified correctly. These models include several predictors such as model type, language, label category, translation status, and if a tweet belongs to train or test set, along with key interaction terms (e.g., Model $\times$ Language, Model $\times$ Label). All models are estimated at the tweet level, and include robust standard errors. Full model formulas, coefficient tables, and additional robustness checks are provided in SI Appendix.

# References

[1] Mens, G. L & Gallego, A. (2025) Positioning political texts with large language models by asking and averaging. *Political Analysis* **0**, 1–9.

[2] Loukas, L, Stogiannidis, I, Diamantopoulos, O, Malakasiotis, P, & Vassos, S. (2023) Making llms worth every penny: Resource-limited text classification in banking. *ICAIF 2023 - 4th ACM International Conference on AI in Finance* **1**, 392–400.

[3] Nasuto, A & Rowe, F. (2024) Understanding anti-immigration sentiment spreading on twitter. *PLOS ONE* **19**, e0307917.

[4] Iacus, S. M, Qi, H, & Han, J. (2025) Deep literature reviews: an application of fine-tuned language models to migration research.

[5] Gilardi, F, Alizadeh, M. I, & Kubli, M. I. (2023) Chatgpt outperforms crowd workers for text-annotation tasks. **120**.

[6] Devlin, J, Chang, M. W, Lee, K, & Toutanova, K. (2018) Bert: Pre-training of deep bidirectional transformers for language understanding. *NAACL HLT 2019 - 2019 Conference of the North American Chapter of the Association for Computational Linguistics: Human Language Technologies - Proceedings of the Conference* **1**, 4171–4186.

[7] Albladi, A, Islam, M, Das, A, Bigonah, M, Zhang, Z, Jamshidi, F, Rahgouy, M, Raychawdhary, N, Marghitu, D, & Seals, C. (2025) Hate speech detection using large language models: A comprehensive review. *IEEE Access* **13**, 20871–20892.

[8] Pires, T, Schlinger, E, & Garrette, D. (2019) How multilingual is multilingual bert? *ACL 2019 - 57th Annual Meeting of the Association for Computational Linguistics, Proceedings of the Conference* pp. 4996–5001.

[9] AI, M. (2024) The llama 3 herd of models.

[10] Wei, J, Tay, Y, Bommasani, R, Raffel, C, Zoph, B, Borgeaud, S, Yogatama, D, Bosma, M, Zhou, D, Metzler, D, Chi, E. H, Hashimoto, T, Vinyals, O, Liang, P, Dean, J, & Fedus, W. (2022) Emergent abilities of large language models. *Transactions on Machine Learning Research*.

[11] Zhang, J, Huang, Y, Liu, S, Gao, Y, & Hu, X. (2025) Do bert-like bidirectional models still perform better on text classification in the era of llms? pp. 18980–18989.

[12] Singh, D. D, Bhattacharjee, R, & Chakraborty, A. (2025) Rethinking hate speech detection on social media: Can llms replace traditional models?

[13] Castillo-López, G, Riabi, A, & Seddah, D. (2023) Analyzing zero-shot transfer scenarios across spanish variants for hate speech detection. pp. 1–13.

[14] Röttger, P, Nozza, D, Bianchi, F, & Hovy, D. (2022) Data-efficient strategies for expanding hate speech detection into under-resourced languages. *Proceedings of the 2022 Conference on Empirical Methods in Natural Language Processing, EMNLP 2022* pp. 5674–5691.

[15] Wang, R, Gong, . Y, Liu, X, Zhao, G, Yang, Z, Guo, B, Zha, Z, & Cheng, P. (2025) Optimizing large language model training using fp4 quantization.

[16] Bizzoni, Y, Juzek, T. S, Na-Bonet, C. E, Chowdhury, K. D, Genabith, J. V, & Teich, E. (year?) How human is machine translationese? comparing human and machine translations of text and speech. pp. 280–290.

[17] Ji, M, Bouillon, P, & Seligman, M. (2023) Cultural and linguistic bias of neural machine translation technology. *Translation Technology in Accessible Health Communication* pp. 100–128.

[18] Savoldi, B, Gaido, M, Bentivogli, L, Negri, M, & Turchi, M. (2021) Gender bias in machine translation. *Transactions of the Association for Computational Linguistics* **9**, 845–874.

[19] Alesina, A & Tabellini, M. (2024) The political effects of immigration: Culture or economics? *Journal of Economic Literature* **62**, 5–46.

[20] Bursztyn, L, Egorov, G, Enikolopov, R, & Petrova, M. (2019) Social media and xenophobia: Evidence from russia, (National Bureau of Economic Research), Technical report.

[21] Liu, Z, Dey, P, tse Huang, J, Zhao, Z, Jiang, B, Gupta, R, Liu, Y, Du, Y, & Zhao, J. (2025) Can llms grasp implicit cultural values? benchmarking llms' cultural intelligence with cq-bench.

[22] Ahuja, K, Diddee, H, Hada, R, Ochieng, M, Ramesh, K, Jain, P, Nambi, A, Ganu, T, Segal, S, Axmed, M, Bali, K, & Sitaram, S. (2023) Mega: Multilingual evaluation of generative ai.

[23] Whitehouse, C, Choudhury, M, & Aji, A. F. (2023) Llm-powered data augmentation for enhanced cross-lingual performance. *EMNLP 2023 - 2023 Conference on Empirical Methods in Natural Language Processing, Proceedings* pp. 671–686.

[24] Lewis, B & Kakkar, D. (2016) Harvard cga geotweet archive v2.0.

[25] Artetxe, M, Goswami, V, Bhosale, S, Fan, A, & Zettlemoyer, L. (2023) Revisiting machine translation for cross-lingual classification. *EMNLP 2023 - 2023 Conference on Empirical Methods in Natural Language Processing, Proceedings* pp. 6489–6499.

[26] Lai, V. D, Ngo, N. T, Veyseh, A. P. B, Man, H, Dernoncourt, F, Bui, T, & Nguyen, T. H. (2023) Chatgpt beyond english: Towards a comprehensive evaluation of large language models in multilingual learning. *Findings of the Association for Computational Linguistics: EMNLP 2023* pp. 13171–13189.

[27] Zhang, S, Gao, C, Zhu, W, Chen, J, Huang, X, Han, X, Feng, J, Deng, C, & Huang, S. (2024) Getting more from less: Large language models are good spontaneous multilingual learners. *EMNLP 2024 - 2024 Conference on Empirical Methods in Natural Language Processing, Proceedings of the Conference* pp. 8037–8051.

[28] Carammia, M, Iacus, S. M, & Porro, G. (2024) Rethinking scale: The efficacy of fine-tuned open-source llms in large-scale reproducible social science research.

[29] Hu, E, Shen, Y, Wallis, P, Allen-Zhu, Z, Li, Y, Wang, S, Wang, L, & Chen, W. (2021) Lora: Low-rank adaptation of large language models. *ICLR 2022 - 10th International Conference on Learning Representations.*

[30] Zhao, Y, Du, L, Ding, X, Xiong, K, Sun, Z, Shi, J, Liu, T, & Qin, B. (year?) Deciphering the impact of pretraining data on large language models through machine unlearning. pp. 9386–9406.

[31] Alkhamissi, B, Elnokrashy, M, Alkhamissi, M, & Diab, M. (year?) Investigating cultural alignment of large language models.

[32] Huo, W, Feng, X, Huang, Y, Fu, C, Li, B, Ye, Y, Zhang, Z, Tu, D, Tang, D, Lu, Y, Wang, H, & Qin, B. (2025) Enhancing non-english capabilities of english-centric large language models through deep supervision fine-tuning.

[33] Guo, Y, Conia, S, Zhou, Z, Li, M, Potdar, S, & Xiao, H. (year?) Do large language models have an english "accent"? evaluating and improving the naturalness of multilingual llms.

[34] Imran, M, Qazi, U, & Ofli, F. (2022) Tbcov: Two billion multilingual covid-19 tweets with sentiment, entity, geo, and gender labels. *Data* **7**.

[35] Jegham, N, Abdelatti, M, Koh, C. Y, Elmoubarki, L, & Hendawi, A. (2025) How hungry is ai? benchmarking energy, water, and carbon footprint of llm inference.

# SI Appendix

Table 1: Twitter/X Search Terms by Language

| Language | Search Term |
|---|---|
| English | immigration, migration, refugee, migrant, illegal immigration,<br>border control, migration crisis, foreigner, asylum seeker, asylee<br>#refugee, #migrant, #immigration |
| Spanish | migrante, refugiad, inmigrant, migración, inmigración<br>#inmigracion, #migrantes |
| Arabic | الهجرة, أزمة الحدود, مراقبة شرعية, غير هجرة مهاجر, لاجئ, هجرة,<br>لجوء طالب أجنبي,<br>#هجرة #لاجئون, |
| French | immigration, migrant, immigrant, immigré, réfugié, clandestin,<br>immigrants illégaux, immigrants, migrants<br>#immigration, #migrants |
| German | Einwanderung, Zuwanderung, Migrant, Einwanderer, Zuwanderer,<br>Flüchtling, illegale Einwanderer, Grenzkontrolle,<br>Asylant, Asylbewerber, Asylsuchende, Asylwerber,<br>Menschenhandel, Schleuser, Flüchtlingskrise<br>#Einwanderung, #Migranten |
| Hindi | आव्रजन, प्रवास, शरणार्थी, प्रवासी, अवैध आव्रजन,<br>सीमा नियंत्रण, प्रवास संकट, विदेशी, शरणार्थी आवेदनकर्ता<br>#शरणार्थी, #आव्रजन |
| Hungarian | bevándorlás, migráció, menekült, migráns, illegális bevándorlás,<br>határellenőrzés, migrációs válság, külföldi, menedékkérő,<br>menedékjogot kérő, bevándorló<br>#menekültek, #bevándorlás |
| Indonesian | imigrasi, migrasi, pengungsi, migran, imigrasi ilegal,<br>kontrol perbatasan, krisis migrasi, orang asing, pencari suaka<br>#pengungsi, #imigrasi |
| Italian | immigra, migrant, rifugiat, clandestin, invasione, sbarchi,<br>immigrazione clandestina<br>#immigrazione, #migranti |
| Polish | imigracja, migracja, uchodźc, migrant, nielegalna imigracja,<br>kontrola graniczna, kryzys migracyjny, cudzoziemiec,<br>osoba ubiegająca się o azyl, azylant<br>#uchodźcy, #imigracja |
| Portuguese | Imigração, migração, migrante, imigrante, refugiado, asilad<br>#imigração, #migrantes |
| Turkish | göç, göçmen, mülteci, mülteciler, yasa dışı göçmenler,<br>göçmenler<br>#göç, #göçmenler |
| Korean | 이민, 이주, 난민, 난민들, 이주자, 이주민, 불법 이민,<br>국경 통제, 이주 위기, 외국인, 망명 신청자, 망명자,<br>탈북자, 북한이탈주민, 탈북민, 새터민<br>#난민, #이민 |

Table 2: Number of labeled tweet observations by language used for training, test, and additional test evaluation.

| Language | Training (75%) | Test (25%) | Additional Test |
|---|---|---|---|
| English | 1125 | 375 | 0 |
| Spanish | 750 | 250 | 0 |
| Turkish | 75 | 25 | 400 |
| Arabic | 75 | 25 | 200 |
| Italian | 75 | 25 | 100 |
| Indonesian | 75 | 25 | 0 |
| Hindi | 75 | 25 | 0 |
| Hungarian | 75 | 25 | 0 |
| Polish | 75 | 25 | 0 |
| Portuguese | 75 | 25 | 100 |
| French | 75 | 25 | 100 |
| German | 75 | 25 | 100 |
| Korean | 0 | 0 | 100 |
| **Total** | **2,700** | **1,100** | **1,100** |

Table 3: Training and test set sizes for each fine-tuned model.

| Model | Training Set Size | Test Set Size |
|---|---|---|
| English-only | 1,125 | 375 |
| Spanish-only | 750 | 250 |
| English–Spanish | 1,875 | 625 |
| Multilanguage (12 lang) | 2,625 | 875 |

Table 4: Language distribution in the LLaMA 2 pretraining data for languages with at least 0.005% share. Most data is in English; the large "unknown" category is partially made up of programming code.

| Language | Percent | Language | Percent |
|---|---|---|---|
| en | 89.70% | uk | 0.07% |
| unknown | 8.38% | ko | 0.06% |
| de | 0.17% | ca | 0.04% |
| fr | 0.16% | sr | 0.04% |
| sv | 0.15% | id | 0.03% |
| zh | 0.13% | cs | 0.03% |
| es | 0.13% | fi | 0.03% |
| ru | 0.13% | hu | 0.03% |
| nl | 0.12% | no | 0.03% |
| it | 0.11% | ro | 0.03% |
| ja | 0.10% | bg | 0.02% |
| pl | 0.09% | da | 0.02% |
| pt | 0.09% | sl | 0.01% |
| vi | 0.08% | hr | 0.01% |

Table 5: **F1 scores by evaluation language and fine tuned model.** Rows report the evaluation language and the total number of labeled tweets used for evaluation (Size). Columns report average F1 scores across labels (pro immigration, anti immigration, neutral, unrelated) for four fine tuned models: English, English Spanish, Spanish, and Multilanguage, where Multilanguage was fine tuned on labeled data from all languages in the dataset. Korean was not included in any training set and was only used for evaluation. Within each language, the best F1 score across models is shown in bold.

| **Language** | **Size** | **Model** | | | |
|---|---|---|---|---|---|
| | | **English** | **English Spanish** | **Multilanguage** | **Spanish** |
| English | 375 | 0.691 | **0.732** | 0.684 | 0.627 |
| Spanish | 250 | 0.714 | 0.779 | **0.791** | 0.765 |
| Turkish | 425 | 0.630 | 0.544 | **0.700** | 0.531 |
| Arabic | 225 | 0.837 | 0.836 | **0.843** | 0.830 |
| Italian | 125 | 0.458 | 0.532 | **0.595** | 0.455 |
| Indonesian | 25 | 0.859 | 0.899 | **0.958** | 0.882 |
| Hindi | 25 | **0.783** | 0.656 | 0.718 | 0.656 |
| Hungarian | 25 | 0.654 | 0.496 | **0.702** | 0.517 |
| Polish | 25 | 0.546 | 0.500 | **0.630** | 0.523 |
| Portuguese | 125 | 0.802 | 0.783 | **0.853** | 0.819 |
| French | 125 | 0.683 | 0.673 | **0.692** | 0.665 |
| German | 25 | 0.269 | **0.518** | 0.372 | 0.384 |
| Korean | 70 | **0.922** | 0.901 | 0.912 | 0.894 |
| Multilingual | 875 | 0.672 | 0.686 | **0.695** | 0.642 |

# Model Specifications

## Modelling the LLM performances

The logistic regression model is expressed as follows:

$$\log\left(\frac{\Pr(\text{Correct}=1)}{\Pr(\text{Correct}=0)}\right) = \beta_0 + \beta_1 \cdot \text{Model} + \beta_2 \cdot \text{Language} + \beta_3 \cdot \text{Train_Test} + \beta_4 \cdot \text{TranslationQuality} + \beta_5 \cdot \text{Label} + \epsilon \quad (1)$$

The variable Correct is a binary indicator equal to 1 if the model's predicted label matches the human-assigned label for the tweet, and 0 otherwise. $\beta_0$ is the intercept, representing the baseline log odds of a correct classification when all predictors are at their reference levels, meaning the model is fine tuned on English data, applied to untranslated tweets labeled as neutral, written in English, and belonging to the training set. $\beta_1$ is the effect of the fine-tuned model type on the log-odds of a correct prediction, $\beta_2$ is the effect of the language in which the tweet is written, and $\beta_3$ is the effect of whether the observation comes from the training or testing set, as indicated by the variable Train/Test. $\beta_4$ is the effect of the presence and quality of translation, captured by TranslationQuality, and $\beta_5$ is the effect of the type of immigration-related label assigned to the tweet (e.g., pro-immigration, anti-immigration, unrelated). Finally, $\epsilon$ is the error term, capturing residual variation not explained by the model.

## Model Specification with Interaction Terms

The logistic regression model with interaction terms between `Model` and `Language` is specified as follows:

$$\log\left(\frac{\Pr(\text{Correct}=1)}{\Pr(\text{Correct}=0)}\right) = \beta_0 + \beta_1 \cdot \text{Model} + \beta_2 \cdot \text{Language} + \beta_3 \cdot \text{Train_Test} + \beta_4 \cdot \text{TranslationQuality} + \beta_5 \cdot \text{Label} + \beta_6 \cdot (\text{Model} \times \text{Language}) + \epsilon \tag{2}$$

The variable `Correct` is a binary indicator that takes the value 1 if the model's predicted label matches the human-assigned label for the tweet, and 0 otherwise. In the model, $\beta_0$ denotes the intercept, representing the baseline log-odds of a correct classification. The coefficient $\beta_1$ captures the main effect of the type of fine-tuned model on the likelihood of a correct prediction, while $\beta_2$ reflects the influence of the tweet's language. The variable `Train/Test`, associated with $\beta_3$, indicates whether the observation comes from the training or test set. Translation quality, coded as `TranslationQuality` and including categories such as Good, Bad, or Unknown, is represented by $\beta_4$. The coefficient $\beta_5$ accounts for the tweet's assigned label category (`Label`), which distinguishes between anti-immigration, pro-immigration, and unrelated content. Importantly, $\beta_6$ captures the interaction between the model and language, allowing for heterogeneity in model performance across languages. Finally, $\epsilon$ denotes the error term, representing variation not explained by the model.

### Model 1: Main Effects Only

Model 1 tests how well different models perform overall and how factors like language, label type, and translation affect classification accuracy. It provides a baseline understanding of the main drivers of model performance. The model estimates the log-odds that a single tweet is correctly classified, based on the type of LLM used (`Model`), the tweet's label as annotated by a human (`Label`; e.g., pro-immigration, anti-immigration, or unrelated), the language in which the tweet is written (`Language`; e.g., Turkish, Spanish), whether the tweet belongs to the training or test split (`Test Set (vs. Train)`), and the quality of translation if the tweet was translated (`TranslationQuality`).

$$\log\left(\frac{\Pr(\text{Correct}=1)}{\Pr(\text{Correct}=0)}\right) = \beta_0 + \beta_1 \cdot \text{Model} + \beta_2 \cdot \text{Label} + \beta_3 \cdot \text{Language} + \beta_4 \cdot \text{Train_Test} + \beta_5 \cdot \text{TranslationQuality} + \epsilon \tag{3}$$

Table 7: **Model 1: Logistic Regression Coefficients (Log-Odds)**

| **Variable** | **Estimate** | **Std. Error** | **p-value** |
|---|---|---|---|
| Intercept | 0.53301 | 0.04129 | < 2e-16 *** |
| Model: English-Spanish | 0.27960 | 0.03840 | < 3.30e-13 *** |
| Model: Multilanguage | 0.26335 | 0.03833 | < 6.37e-12 *** |
| Model: Spanish | -0.21494 | 0.03675 | < 4.98e-8 *** |
| *Label (ref: Neutral)* | | | |
| Label: Unrelated | 1.57870 | 0.03693 | < 2e-16 *** |
| Label: Anti-immigration | 0.21475 | 0.04049 | < 1.13e-7 *** |
| Label: Pro-immigration | 0.22177 | 0.04028 | < 3.69e-8 *** |
| *Language (ref: English)* | | | |
| Language: Arabic | -0.40770 | 0.06479 | < 3.13e-8 *** |
| Language: French | -0.63238 | 0.06418 | < 2e-16 *** |
| Language: German | -0.67239 | 0.07439 | < 2e-16 *** |
| Language: Hindi | -0.38607 | 0.08903 | < 1.45e-4 *** |
| Language: Hungarian | -0.86495 | 0.07861 | < 2e-16 *** |
| Language: Indonesian | 0.04980 | 0.10677 | 0.64089 |
| Language: Italian | -0.70959 | 0.06403 | < 2e-16 *** |
| Language: Korean | 0.59410 | 0.21986 | 0.00689 ** |
| Language: Polish | -1.03426 | 0.07744 | < 2e-16 *** |
| Language: Portuguese | -0.19452 | 0.07591 | 0.01039 * |
| Language: Spanish | -0.32283 | 0.03497 | < 2e-16 *** |
| Language: Turkish | -0.72879 | 0.05235 | < 2e-16 *** |
| *Other Controls* | | | |
| Test Set (vs. Train) | -0.23528 | 0.02881 | < 3.14e-15 *** |
| Translation Quality: Bad | 0.04676 | 0.16514 | 0.77704 |
| Translation Quality: Good | -0.38126 | 0.06473 | < 3.86e-8 *** |
| Translation Quality: Unknown | -0.68046 | 0.14026 | < 1.23e-6 *** |

*Note:* Regression results without interaction terms. Significance codes: *** p<0.001, ** p<0.01, * p<0.05.

### Model 2: Interaction Between Model and Label

Model 2 explores whether some models are better at handling specific types of content (eg. pro- or anti-immigration content). It tests how accuracy changes across the type of fine-tuned LLM depending on the kind of label being predicted. Specifically, Model 2 examines how classification accuracy varies by label category (eg. pro-immigration, anti-immigration) by including interaction terms between `Model` type and `Label`. The model also controls for the language of the tweet via `Language` (e.g., Turkish, Spanish), `Train_Test` (indicating whether the tweet is in the training or test split), and `TranslationQuality` (e.g., good, bad, or unknown).

$$\log\left(\frac{\Pr(\text{Correct}=1)}{\Pr(\text{Correct}=0)}\right) = \beta_0 + \beta_1 \cdot (\text{Model} \times \text{Label}) + \beta_2 \cdot \text{Language} + \beta_3 \cdot \text{Train_Test} + \beta_4 \cdot \text{TranslationQuality} + \epsilon \tag{4}$$

Table 8: **Model 2: Logistic Regression Coefficients (Log-Odds)**

| **Variable** | **Estimate** | **Std. Error** | **p-value** |
|---|---|---|---|
| Intercept | 0.74766 | 0.05769 | < 2e-16 *** |
| *Main Effects* | | | |
| Model: English-Spanish | -0.32094 | 0.07965 | < 5.59e-5 *** |
| Model: Multilanguage | -0.09746 | 0.07995 | 0.22288 |
| Model: Spanish | -0.05920 | 0.08006 | 0.45960 |
| Label: Anti-immigration | 0.55371 | 0.07599 | < 3.18e-13 *** |
| Label: Pro-immigration | 0.06383 | 0.07401 | 0.38845 |
| Label: Unrelated | 1.03457 | 0.06353 | < 2e-16 *** |
| Language: Arabic | -0.44719 | 0.06681 | < 2.18e-10 *** |
| Language: French | -0.68573 | 0.06632 | < 2e-16 *** |
| Language: German | -0.72643 | 0.07629 | < 2e-16 *** |
| Language: Hindi | -0.40824 | 0.09073 | < 6.81e-6 *** |
| Language: Hungarian | -0.92702 | 0.08054 | < 2e-16 *** |
| Language: Indonesian | 0.04013 | 0.10897 | 0.71271 |
| Language: Italian | -0.76066 | 0.06581 | < 2e-16 *** |
| Language: Korean | 0.57407 | 0.22254 | 0.00989 ** |
| Language: Polish | -1.11436 | 0.07995 | < 2e-16 *** |
| Language: Portuguese | -0.21659 | 0.07817 | 0.00559 ** |
| Language: Spanish | -0.34896 | 0.03613 | < 2e-16 *** |
| Language: Turkish | -0.77071 | 0.05377 | < 2e-16 *** |
| Test Set (vs. Train) | -0.24988 | 0.02965 | < 2e-16 *** |
| Translation Quality: Bad | 0.09814 | 0.15943 | 0.53819 |
| Translation Quality: Good | -0.34752 | 0.06269 | < 2.96e-8 *** |
| Translation Quality: Unknown | -0.52834 | 0.13596 | < 1.02e-4 *** |
| *Interaction Effects (Model × Label)* | | | |
| English-Spanish × Anti-immigration | -0.45531 | 0.11133 | < 4.32e-5 *** |
| Multilanguage × Anti-immigration | 0.19795 | 0.11451 | 0.08386 . |
| Spanish × Anti-immigration | -1.09440 | 0.11197 | < 2e-16 *** |
| English-Spanish × Pro-immigration | 0.66239 | 0.11167 | < 2.99e-6 *** |
| Multilanguage × Pro-immigration | 1.62200 | 0.12667 | < 2e-16 *** |
| Spanish × Pro-immigration | -1.19599 | 0.11128 | < 2e-16 *** |
| English-Spanish × Share Unrelated | 1.75508 | 0.10860 | < 2e-16 *** |
| Multilanguage × Share Unrelated | 0.11917 | 0.09469 | 0.20819 |
| Spanish × Share Unrelated | 0.93977 | 0.10114 | < 2e-16 *** |

*Note:* Logistic regression results estimating the probability of correct classification based on model, label, and their interaction. Significance codes: *** p<0.001, ** p<0.01, * p<0.05, . p<0.1.

### Model 3: Interaction Between Model and Language

Model 3 examines whether LLM accuracy changes across languages (eg. Spanish, German) of the humanly annotead tweets. In details, Model 3 includes an interaction term between `Model` and `Language` and it expands the baseline Model 1 by retaining all other variables from Model 1.

$$\log\left(\frac{\Pr(\texttt{Correct}=1)}{\Pr(\texttt{Correct}=0)}\right) = \beta_0 + \beta_1 \cdot (\texttt{Model} \times \texttt{Language}) + \beta_2 \cdot \texttt{Label} + \beta_3 \cdot \texttt{Train_Test} + \beta_4 \cdot \texttt{TranslationQuality} + \epsilon \tag{5}$$

Table 9: **Model 3: Logistic Regression Coefficients (Log-Odds)**

| **Variable** | **Estimate** | **Std. Error** | **p-value** |
|---|---|---|---|
| Intercept | 0.83547 | 0.05407 | < 2e-16 *** |
| *Main Effects* | | | |
| Model: English-Spanish | 0.24164 | 0.06933 | 0.00049 *** |
| Model: Multilanguage | -0.16966 | 0.06560 | 0.00971 ** |
| Model: Spanish | -0.87078 | 0.06220 | < 2e-16 *** |
| Language: Arabic | -0.71633 | 0.11853 | < 1.51e-8 *** |
| Language: French | -0.97501 | 0.11759 | < 2e-16 *** |
| Language: German | -1.03035 | 0.13179 | < 5.36e-15 *** |
| Language: Hindi | -0.57349 | 0.15825 | 0.00029 *** |
| Language: Hungarian | -1.39791 | 0.14315 | < 2e-16 *** |
| Language: Indonesian | -0.74985 | 0.16431 | < 5.02e-6 *** |
| Language: Italian | -1.15841 | 0.11738 | < 2e-16 *** |
| Language: Korean | 0.34114 | 0.43740 | 0.43543 |
| Language: Polish | -1.35079 | 0.14297 | < 2e-16 *** |
| Language: Portuguese | -0.74470 | 0.13037 | < 1.12e-7 *** |
| Language: Spanish | -0.94583 | 0.06693 | < 2e-16 *** |
| Language: Turkish | -0.96396 | 0.09761 | < 2e-16 *** |
| Label: Unrelated | 1.59632 | 0.03719 | < 2e-16 *** |
| Label: Anti-immigration | 0.21947 | 0.04079 | < 7.44e-8 *** |
| Label: Pro-immigration | 0.22459 | 0.04063 | < 3.24e-8 *** |
| Test Set (vs. Train) | -0.23402 | 0.02899 | < 7e-15 *** |
| Translation Quality: Bad | 0.18462 | 0.16596 | 0.26593 |
| Translation Quality: Good | -0.17703 | 0.06793 | 0.00916 ** |
| Translation Quality: Unknown | -0.46301 | 0.15715 | 0.00322 ** |
| *Interaction Effects (Model × Language)* | | | |
| English-Spanish × Arabic | -0.13626 | 0.17810 | 0.44422 |
| Multilanguage × Arabic | 0.35275 | 0.17829 | 0.04787 * |
| Spanish × Arabic | 0.80012 | 0.17218 | < 3.37e-6 *** |
| English-Spanish × French | -0.22181 | 0.17753 | 0.21149 |
| Multilanguage × French | 0.59752 | 0.17971 | 0.00088 *** |
| Spanish × French | 0.83021 | 0.17457 | < 1.98e-6 *** |
| English-Spanish × German | -0.18031 | 0.20483 | 0.37872 |
| Multilanguage × German | 0.73363 | 0.20859 | 0.00044 *** |
| Spanish × German | 0.71269 | 0.20217 | 0.00042 *** |
| English-Spanish × Hindi | -0.43905 | 0.24097 | 0.06846 . |
| Multilanguage × Hindi | 0.03451 | 0.24187 | 0.88654 |
| Spanish × Hindi | 0.79986 | 0.24310 | 0.00100 ** |
| English-Spanish × Hungarian | -0.05599 | 0.21602 | 0.79546 |
| Multilanguage × Hungarian | 1.04560 | 0.22558 | < 3.57e-6 *** |
| Spanish × Hungarian | 1.05643 | 0.21385 | < 7.81e-7 *** |
| English-Spanish × Indonesian | 0.94326 | 0.32887 | 0.00413 ** |
| Multilanguage × Indonesian | 1.13799 | 0.30831 | 0.00022 *** |
| Spanish × Indonesian | 1.44188 | 0.27838 | < 2.22e-7 *** |
| English-Spanish × Italian | -0.06897 | 0.17688 | 0.69660 |
| Multilanguage × Italian | 0.77518 | 0.17862 | < 1.43e-5 *** |
| Spanish × Italian | 0.96861 | 0.17401 | < 2.60e-6 *** |
| English-Spanish × Korean | -0.03758 | 0.64417 | 0.95348 |
| Multilanguage × Korean | -0.00619 | 0.59769 | 0.99173 |
| Spanish × Korean | 0.87078 | 0.61696 | 0.15812 |
| English-Spanish × Polish | -0.33593 | 0.21419 | 0.11679 |
| Multilanguage × Polish | 0.80255 | 0.22109 | 0.00028 *** |
| Spanish × Polish | 0.59402 | 0.21151 | 0.00498 ** |
| English-Spanish × Portuguese | 0.19368 | 0.20851 | 0.35296 |
| Multilanguage × Portuguese | 0.89188 | 0.21770 | < 4.19e-5 *** |
| Spanish × Portuguese | 1.03750 | 0.19858 | < 1.75e-7 *** |
| English-Spanish × Spanish | 0.43262 | 0.10153 | < 2.03e-5 *** |
| Multilanguage × Spanish | 0.57226 | 0.09775 | < 6.10e-9 *** |
| Spanish × Spanish | 1.34904 | 0.09538 | < 2e-16 *** |
| English-Spanish × Turkish | -0.40753 | 0.14027 | 0.00367 ** |
| Multilanguage × Turkish | 0.56074 | 0.14251 | < 8.33e-5 *** |
| Spanish × Turkish | 0.62690 | 0.13657 | < 4.42e-6 *** |

*Note:* Significance codes: *** p<0.001, ** p<0.01, * p<0.05, . p<0.1.

### Model 4: Interaction Between Model and Share of Unrelated Content

Model 4 investigates how the presence of unrelated content to immigration affects LLMs accuracy. This model extends the baseline Model 1 by including the interaction terms between model type and the share of unrelated tweets.

$$\log\left(\frac{\Pr(\text{Correct}=1)}{\Pr(\text{Correct}=0)}\right) = \beta_0 + \beta_1 \cdot (\text{Model} \times \text{Share_Unrelated}) + \beta_2 \cdot (\text{Model} \times \text{Language}) + \beta_3 \cdot \text{Language} + \beta_4 \cdot \text{Train_Test} + \beta_5 \cdot \text{TranslationQuality} + \epsilon \quad (6)$$

Table 10: **Model 4 – Logistic Regression Coefficients (Log-Odds)**

| **Variable** | **Estimate** | **Std. Error** | **p-value** |
|---|---|---|---|
| (Intercept) | -0.556 | 0.291 | 0.056 . |
| *Main Effects* | | | |
| Model: English–Spanish | 0.420 | 0.465 | 0.366 |
| Model: Multilanguage | 1.020 | 0.477 | 0.032 * |
| Model: Spanish | -1.052 | 0.454 | 0.021 * |
| Language: Arabic | -1.475 | 0.235 | <0.001 *** |
| Language: French | -0.608 | 0.148 | <0.001 *** |
| Language: German | -0.290 | 0.201 | 0.149 |
| Language: Hindi | -1.598 | 0.304 | <0.001 *** |
| Language: Hungarian | -1.586 | 0.188 | <0.001 *** |
| Language: Indonesian | -2.231 | 0.378 | <0.001 *** |
| Language: Italian | -0.673 | 0.159 | <0.001 *** |
| Language: Korean | -1.200 | 0.541 | 0.026 * |
| Language: Polish | -1.438 | 0.187 | <0.001 *** |
| Language: Portuguese | -1.738 | 0.248 | <0.001 *** |
| Language: Spanish | -1.330 | 0.135 | <0.001 *** |
| Language: Turkish | -1.149 | 0.125 | <0.001 *** |
| Share Unrelated | 4.661 | 0.665 | <0.001 *** |
| Test Set (vs. Train) | -0.216 | 0.040 | <0.001 *** |
| Translation Quality: Bad | 0.310 | 0.162 | 0.055 . |
| Translation Quality: Good | -0.183 | 0.073 | 0.012 * |
| Translation Quality: Unknown | -0.415 | 0.173 | 0.016 * |
| *Model × Language Interactions* | | | |
| English–Spanish × Arabic | -0.001 | 0.392 | 0.999 |
| Multilanguage × Arabic | 1.133 | 0.409 | 0.006 ** |
| Spanish × Arabic | 0.522 | 0.380 | 0.169 |
| English–Spanish × French | -0.222 | 0.225 | 0.322 |
| Multilanguage × French | 0.307 | 0.225 | 0.173 |
| Spanish × French | 0.821 | 0.221 | 0.000 *** |
| English–Spanish × German | -0.294 | 0.332 | 0.377 |
| Multilanguage × German | 0.129 | 0.341 | 0.706 |
| Spanish × German | 0.729 | 0.326 | 0.025 * |
| English–Spanish × Hindi | -0.298 | 0.486 | 0.539 |
| Multilanguage × Hindi | 0.959 | 0.495 | 0.053 . |
| Spanish × Hindi | 0.484 | 0.482 | 0.316 |
| English–Spanish × Hungarian | -0.017 | 0.296 | 0.954 |
| Multilanguage × Hungarian | 1.174 | 0.310 | 0.000 *** |
| Spanish × Hungarian | 0.916 | 0.292 | 0.002 ** |
| English–Spanish × Indonesian | 1.161 | 0.676 | 0.086 . |
| Multilanguage × Indonesian | 2.518 | 0.673 | 0.000 *** |
| Spanish × Indonesian | 1.095 | 0.621 | 0.078 . |
| English–Spanish × Italian | -0.066 | 0.250 | 0.792 |
| Multilanguage × Italian | 0.397 | 0.254 | 0.118 |
| Spanish × Italian | 0.978 | 0.244 | 0.000 *** |
| English–Spanish × Korean | 0.178 | 0.826 | 0.829 |
| Multilanguage × Korean | 1.351 | 0.801 | 0.092 . |
| Spanish × Korean | 0.491 | 0.798 | 0.538 |
| English–Spanish × Polish | -0.328 | 0.291 | 0.259 |
| Multilanguage × Polish | 0.897 | 0.302 | 0.003 ** |
| Spanish × Polish | 0.455 | 0.286 | 0.112 |
| English–Spanish × Portuguese | 0.377 | 0.413 | 0.362 |
| Multilanguage × Portuguese | 1.709 | 0.434 | 0.000 *** |
| Spanish × Portuguese | 0.905 | 0.400 | 0.023 * |
| English–Spanish × Spanish | 0.431 | 0.209 | 0.039 * |
| Multilanguage × Spanish | 0.994 | 0.209 | 0.000 *** |
| Spanish × Spanish | 1.097 | 0.200 | 0.000 *** |
| English–Spanish × Turkish | -0.367 | 0.188 | 0.051 . |
| Multilanguage × Turkish | 0.674 | 0.190 | 0.000 *** |
| Spanish × Turkish | 0.427 | 0.182 | 0.019 * |
| *Model × Share Unrelated Interactions* | | | |
| English–Spanish × Share Unrelated | -0.431 | 1.056 | 0.683 |
| Multilanguage × Share Unrelated | -2.740 | 1.088 | 0.012 * |
| Spanish × Share Unrelated | 0.605 | 1.036 | 0.559 |

*Note:* Logistic regression results from Model 4, including interaction terms between model type, language, and the share of unrelated tweets. Significance codes: *** p<0.001, ** p<0.01, * p<0.05, . p<0.1.

Table 6: Model type, parameter scale, inference speed, water usage, $CO_2$ emissions, and cost to classify 10 billion tweets

| Provider | Model | Parameters | Speed† (token per seconds) | Energy (MWh) | Water ($m^3$) | $CO_2$ (t) | Cost (USD) |
|---|---|---|---|---|---|---|---|
| **Meta** | **Llama 3.2 - FT** | **3B** | **3,854.4**‡ | **28.5** | **14.3** | **0** | **$ 2.9 k** |
| Meta | Llama 3.2 | 3B | 107* | 1026.7 | 513.4 | 0 | — |
| Meta | Llama 3.3 | 70B | 87* | 1262.8 | 631.4 | 0 | — |
| Meta | Llama 4 “Maverick” | 405B | 158* | 695.3 | 347.7 | 0 | — |
| OpenAI | GPT-3.5 Turbo | 175B | 104* | 1056.4 | 3636.0 | 373 | $ 0.580 M |
| OpenAI | GPT-4.1 | 1.7T | 143* | 768.2 | 2644.3 | 271 | $ 3.48 M |
| OpenAI | GPT-5.2 pro | 1.7T | — | — | — | — | $ 25.20 M |
| Anthropic | Claude Haiku 3.5 | — | 79* | 1390.6 | 4619.7 | 535 | $ 0.936 M |
| Anthropic | Claude Sonnet 4 | — | 68* | 1615.6 | 5367.0 | 622 | $ 3.51 M |
| Anthropic | Claude Opus 4 | — | 70* | 1569.4 | 5213.7 | 604 | $ 17.55 M |
| DeepSeek | DeepSeek Chat (V3) | 671B | 27* | 5231.5 | 37750.4 | 3139 | $ 0.316 M |
| DeepSeek | DeepSeek Reasoner (R1) | 671B | 23* | 4776.6 | 34467.7 | 2866 | $ 0.643 M |
| xAI | Grok 4 | 1.7T | 76* | 1445.5 | 4975.6 | 510 | $ 3.54 M |

**Notes:** Estimates assume classifying 10 billion tweets. Each tweet uses a 76-token prompt plus an average tweet length of 36 tokens per tweet, totalling 112 input tokens while output token is just 1; overall inference therefore processes 1.12 trillion input tokens and 10 billion output tokens. Fine-tuning is simulated across four models trained on a combined 229 500 tokens. In bold on top, Llama 3.2 - FT is one of our fine-tuned model based on Llama 3.2 - 3B open-source model developed by Meta. **Parameters**: Parameter ranges marked * are non-official estimates: GPT-3.5 Turbo (OpenGenus — 175 B*), GPT-4o (SemiAnalysis 2025 — 1.7–2.0 T*), and Grok 4 1.7 T* (axion.pm). **Speed**: Tokens-per-second (TPS) for our fine-tuned Llama 3.2 3B model comes from our own benchmark on a single NVIDIA A100-80 GB running 4-bit quantisation, whereas all other TPS figures are taken from the throughput estimates published on *artificialanalysis.com*. **Energy** (MWh) follows $\text{Energy} = \frac{\text{tokens}}{\text{TPS}} \times \frac{0.35\ \text{kW}}{3600}$ with a 0.35 kW average GPU+host draw. The constant 0.35 kW is derived from the NVIDIA A100-80 GB PCIe product brief’s 300 W board TDP (NVIDIA, 2023) plus 50 W of host-side overhead for CPU, memory, networking, cooling, and PSU losses observed in single-GPU inference servers, giving 350 W total. **Water** ($m^3$) uses Eq. 3 of Jegham *et al.* (2025) [35]: $\text{Water} = \text{Energy}\,(\text{WUE}_{\text{site}} + \text{WUE}_{\text{source}})$, where $WUE$ is water-usage effectiveness in litres $\text{kWh}^{-1}$ and is set at 3.442 for Azure, 3.322 for AWS, 7.216 for DeepSeek, and capped at 0.5 for MGHPCC (the Massachusetts Green High-Performance Computing Center, a hydro-powered academic datacentre in Holyoke, MA). **$CO_2$**: Carbon emissions apply Eq. 4 of Jegham *et al.* (2025) [35]: $CO_2 = \text{Energy} \times \text{CIF}$, where CIF (Carbon Intensity Factor) the average life-cycle $CO_2$ released per kilowatt-hour of electricity consumed (kg $\text{kWh}^{-1}$) for a provider’s data-centre supply. We use CIF values of 0.3528 (Azure), 0.385 (AWS) and 0.600 (DeepSeek); MGHPCC’s $\geq$90 % hydro grid gives an effective CIF near zero, so our fine-tuned model runs are treated as carbon-free. **Cost**: we use the following input/output rates (USD per million tokens): GPT-3.5 0.50/1.50, GPT-4o 2.50/10, GPT-5.2 1.750/14 (cached input 0.175), Haiku 0.80/4, Sonnet 3/15, Opus 15/75, DeepSeek Chat 0.27/1.10, Reasoner 0.55/2.19, Grok 4 3/15; thus Claude Opus 4 inference costs $1.13 \times 10^6 \times \$15 + 10^4 \times \$75 = \$17.7$ M, whereas self-hosted Llama 3.2 3B incurs only electricity: 28.5 MWh $\times$\$0.10=\$2 850. All prices reflect August 2025 public API rates per million tokens. All numeric values are rounded to one decimal place (speed, energy, water) or to the nearest tonne / \$1 000 ($CO_2$, cost).